\documentclass[letterpaper, 10 pt, conference]{ieeeconf}  

\IEEEoverridecommandlockouts                              

\usepackage{graphics} 
\usepackage{epsfig} 
\usepackage{times} 
\usepackage{amsmath} 
\usepackage{amssymb}  
\usepackage{cite}
\usepackage{bm}
\usepackage{algorithm}
\usepackage{algpseudocode}
\usepackage{hyperref}
\usepackage{makecell}
\def\thickhline{\noalign{\hrule height.8pt}}
\usepackage{subfigure}
\usepackage{xcolor, soul}
\usepackage{dblfloatfix}  
\usepackage{multirow}
\usepackage{nameref}
\definecolor{gray}{rgb}{0.95, 0.95, 0.95}
\definecolor{green}{rgb}{0.9, 0.94, 0.86}
\definecolor{blue}{rgb}{0.86, 0.89, 0.945}
\DeclareMathOperator*{\argmax}{arg\,max}

\title{\LARGE \bf
E2Map: Experience-and-Emotion Map \\for Self-Reflective Robot Navigation with Language Models
}

\author{Chan Kim$^{*1}$, Keonwoo Kim$^{*1}$, Mintaek Oh$^{1}$, Hanbi Baek$^{1}$, Jiyang Lee$^{1}$,\\ Donghwi Jung$^{1}$, Soojin Woo$^{1}$, Younkyung Woo$^{2}$$^{3}$, John Tucker$^{4}$, Roya Firoozi$^{5}$,\\ Seung-Woo Seo$^{1}$, Mac Schwager$^{4}$, and Seong-Woo Kim$^{1}$ 
\thanks{* indicates equal contribution.}
\thanks{$^{1}$ Seoul National University, $^{2}$ Work done while visiting the Autonomous Robot Intelligence Lab, SNU, $^{3}$ Carnegie Mellon University, $^{4}$ Stanford University, $^{5}$ University of Waterloo. Correspondence to: Seong-Woo Kim  {\tt\small snwoo@snu.ac.kr}}%
}

\begin{document}

\maketitle
\thispagestyle{empty}
\pagestyle{empty}

\begin{abstract}
Large language models (LLMs) have shown significant potential in guiding embodied agents to execute language instructions across a range of tasks, including robotic manipulation and navigation. However, existing methods are primarily designed for static environments and do not leverage the agent's own experiences to refine its initial plans. Given that real-world environments are inherently stochastic, initial plans based solely on LLMs' general knowledge may fail to achieve their objectives, unlike in static scenarios. To address this limitation, this study introduces the Experience-and-Emotion Map (E2Map), which integrates not only LLM knowledge but also the agent's real-world experiences, drawing inspiration from human emotional responses. The proposed methodology enables one-shot behavior adjustments by updating the E2Map based on the agent's experiences. Our evaluation in stochastic navigation environments, including both simulations and real-world scenarios, demonstrates that the proposed method significantly enhances performance in stochastic environments compared to existing LLM-based approaches. Code and supplementary materials are available at \href{https://e2map.github.io/}{https://e2map.github.io/}.
\end{abstract}

\section{Introduction}

Large language models (LLMs), pre-trained on Internet-scale data, have emerged as a promising method to encapsulate the world's knowledge distilled in language. These LLMs have demonstrated a variety of capabilities, including interpreting and responding to natural language instructions, performing logical reasoning, and generating code. Leveraging these capabilities, many studies have explored applying the generalizable knowledge of LLMs to embodied agents, enabling them to interact physically in the real world.

To effectively utilize LLMs' knowledge in enabling embodied agents, it is crucial to ground this knowledge in the real-world environments where the agents operate. Previous studies have proposed methods to decompose language instructions into sequential subtasks, which are then executed using predefined motion primitives \cite{huang2022language, ahn2022can, huang2022inner}. Other research has explored the use of LLMs' code generation capabilities to translate language instructions into executable code for robots, which is then used in conjunction with various APIs \cite{liang2023code, mu2024robocodex, chen2024roboscript}. Additionally, some studies have focused on grounding information from LLMs and vision-language models (VLMs) in the spatial contexts where robots operate \cite{huang2023voxposer, chen2023open, huang2023visual}. For instance, \cite{huang2023voxposer} proposed a method for planning by representing language instructions as spatial costs and affordances using LLMs' code-writing capabilities. Furthermore, \cite{chen2023open} and \cite{huang2023visual} introduced methodologies for robot navigation that use maps that allow spatial information queries by language.

\begin{figure}[t]
    \centering
    \includegraphics[width=\linewidth]{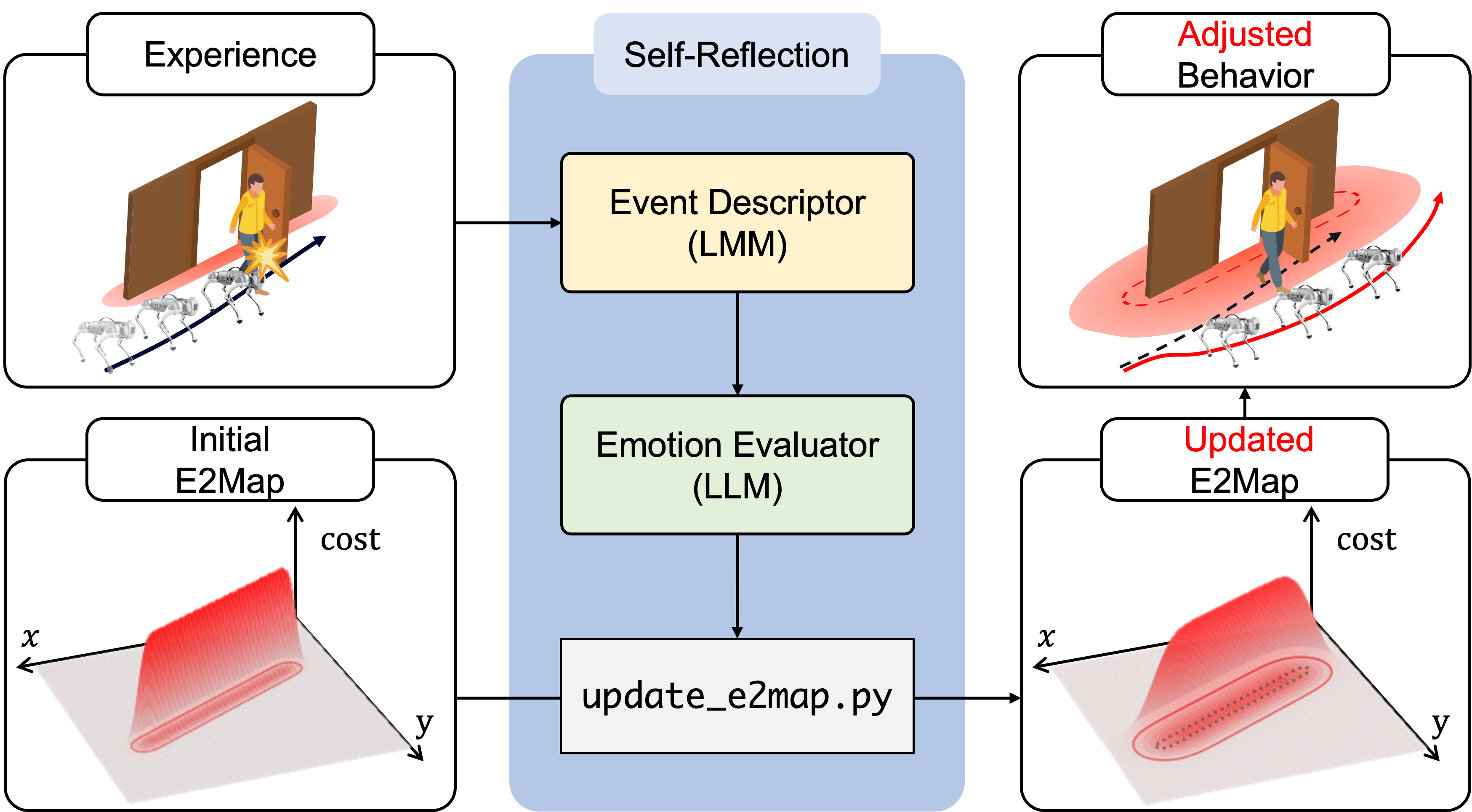}\\[-1.1ex]
    \caption{E2Map is a spatial map that captures the agent's emotional responses to its experiences. Our method enables one-shot behavior adjustments in stochastic environments by updating the E2Map through the diverse capabilities of LLMs and LMMs. Since VLM can refer to both the Vision-Language Model and Visual Language Map, we will preferably use LMM for the former and VLMap for the latter.} \vspace{-1.4em}
\label{fig_main}
\end{figure}

While these methodologies have successfully demonstrated the use of LLMs with embodied agents, they primarily address static environments and fail to incorporate the agent's own experiences. Given that the real world is inherently stochastic and subject to various uncertainties, initial plans based solely on the general knowledge of LLMs may fall short of achieving their objectives. For example, a robot navigating indoors might unexpectedly collide with a person stepping out from a suddenly opened door, potentially causing it to fall. In such cases, the robot may fail to complete its mission using its initial plan. To increase the success rate in similar future scenarios, it is essential to incorporate these experiences and adjust the robot’s behavior to maintain a safe distance when encountering doors. In summary, to achieve objectives in stochastic environments, the robot must refine its plan based on experience.

Humans experience emotions in response to stimuli, which influence specific behaviors toward certain objects or situations. According to \cite{Damasio2021} and \cite{tierney2019power}, emotions play a crucial role in maintaining homeostasis by guiding humans to avoid previously encountered dangers, thus supporting survival. Furthermore, recent research \cite{galvez2021emotional} demonstrates that humans integrate emotional experiences into spatial representations for navigation. By using emotions as spatial boundaries to maintain homeostasis, humans can prefer or avoid specific locations based on their experiences during navigation.

Inspired by human emotional mechanisms, this study proposes a language-based robot system that utilizes the capabilities of LLMs and large multi-modal models (LMMs) to enable an embodied agent to autonomously adjust its behavior in a one-shot manner. The core component is the Experience-and-Emotion Map (E2Map), which serves as a spatial map that grounds both the general knowledge of LLMs and the agent's emotional responses to experiences. When the agent interacts with a particular object or space, it quantifies its emotional response and integrates this information into the E2Map as shown in Fig. \ref{fig_main}. The agent's planning and control module then uses the E2Map as a cost function to guide its actions, allowing for behavior refinement through E2Map updates. The contributions are summarized as follows:

\begin{itemize}
\item Inspired by human emotional mechanisms, we propose E2Map, a spatial representation that integrates both the general knowledge of LLMs and the agent's emotional responses to its experiences to facilitate planning.
\item We propose a robot system that leverages the E2Map and various LLM capabilities—including code generation, event description, and emotion evaluation—to enable the agent to autonomously adjust its behavior in a one-shot manner based on its experiences.
\item We verified that the proposed methodology can autonomously adjust behavior by incorporating its experiences in navigation environments with changing conditions and dynamic objects, demonstrating superior performance over existing LLM-based approaches.
\end{itemize}

\section{Related Works}

\subsection{LLM-based Robotics}

Research on using language instructions to control robots has advanced significantly with the development of LLMs. Numerous studies \cite{huang2022language, ahn2022can, huang2022inner, zeng2022socratic, raman2022planning, song2023llm, ding2023task, liang2023code, mu2024robocodex, chen2024roboscript, huang2023voxposer, yu2023language} have aimed to leverage the general knowledge embedded in LLMs and the benefits of few-shot prompting, which eliminates the need for additional training. \cite{huang2022language, ahn2022can, huang2022inner} proposed a method that uses LLMs to decompose a given language instruction into a sequence of subtasks and employs predefined motion primitives to execute each subtask. \cite{liang2023code, mu2024robocodex, chen2024roboscript} utilized LLMs' capability of code generation to transform language instructions into executable code for robots, using various APIs to carry out tasks. \cite{huang2023voxposer} proposed utilizing LLMs to construct affordances and constraints within a 3D voxel map and employing model predictive control (MPC) to compute actions, eliminating the need for predefined motion primitives.

While these approaches have shown notable success in language-based robotics, they primarily focus on static environments and overlook the integration of the robot's experiences. \cite{yu2023language} demonstrated that it is possible to refine a robot's behavior through additional human language instructions, but this refinement does not stem from the robot's own experiences. In contrast, the proposed methodology addresses behavior refinement in stochastic environments by grounding the robot's emotional responses to its own experiences.
\renewcommand{\arraystretch}{1.2}
\begin{table}[t]
\centering
\caption{Comparison of Existing and Proposed Methods.} \vspace{-1.0em}
\label{table_comparison}
\resizebox{1.0\linewidth}{!}{
\begin{tabular}{c|cccc}
 Methods & LLM-Based & \makecell{Spatial \\ Grounding} & \makecell{Behavior\\ Adjustment} & Self-Reflection \\ \hline \hline
\cite{huang2022language, ahn2022can, huang2022inner, liang2023code, mu2024robocodex, chen2024roboscript, gadre2023cows} & \checkmark &  &  &  \\ \hline
 \cite{huang2023voxposer, chen2023open, huang2023visual} & \checkmark & \checkmark &  &  \\ \hline
 \cite{yu2023language} & \checkmark &  & \checkmark &  \\ \thickhline
 \textbf{Ours} & \checkmark & \checkmark & \checkmark & \checkmark \\ \hline 
\end{tabular}
}
\vspace{-1.5em}
\end{table}

\subsection{Visual-Language Navigation}

To demonstrate the ability to achieve one-shot behavior adjustment in stochastic environments, we evaluated the proposed methodology in a mobile robot indoor navigation scenario. Navigation environments involve dynamic objects such as people, which can cause unexpected events, e.g., a person suddenly stepping out of the door.

Visual-language navigation (VLN), which uses a robot's visual input to execute given language instructions, has been extensively researched to date. Early research used methods to map language instructions and visual input to discrete \cite{anderson2018vision, guhur2021airbert}, or continuous actions \cite{krantz2020beyond, hong2022bridging}. However, these methods are data-intensive, which raises concerns about the cost of data collection. To address these problems, research has emerged that uses foundation models \cite{gadre2023cows, shah2023lm, chen2023open, huang2023visual}. \cite{shah2023lm} proposed a topological graph-based navigation approach using three different foundation models. \cite{gadre2023cows} demonstrated object navigation using open-vocabulary object detector based on CLIP \cite{radford2021learning} and exploration techniques. \cite{chen2023open} proposed creating a prebuilt spatial map that stores visual-language features of objects, while \cite{huang2023visual} suggested generating maps by storing pixel-level visual-language features of spaces. By grounding language instructions into these maps, both approaches demonstrated their effectiveness in enabling spatial open-vocabulary navigation tasks. However, these approaches are limited to static environments and do not account for the refinement of navigation plans based on the agent's real-time experiences. Table \ref{table_comparison} presents a comparison between the proposed method and existing methods.

\begin{figure*}[t]
    \centering
    \includegraphics[width=\linewidth]{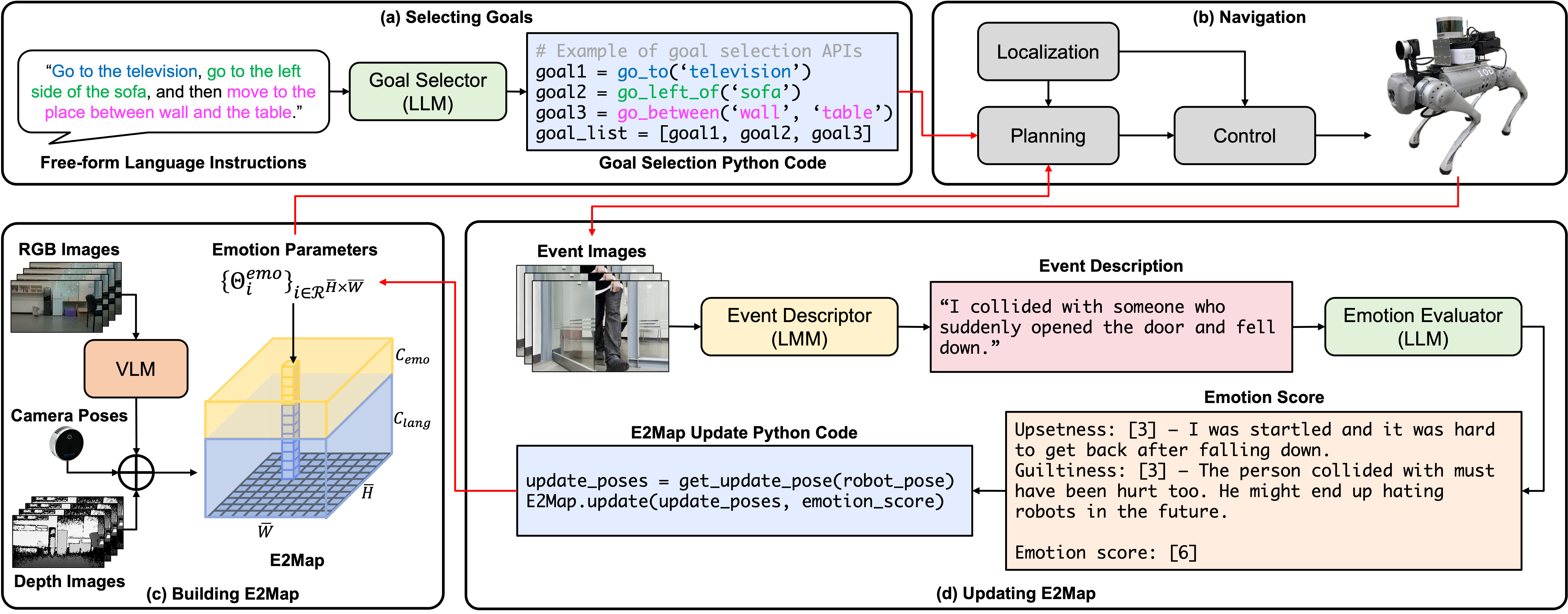}\\[-1.1ex]
    \caption{System Overview: (a) E2Map is created by embedding visual-language features and emotion parameters into corresponding grid cells. (b) When a user provides a language instruction, the \textit{goal selector} generates code through goal selection APIs to decide goals to reach. (c) The planning algorithm then uses emotion as a cost to generate the optimal path to the goal. (d) If the agent encounters unexpected events during navigation, the E2Map is updated through the sequential operation of the \textit{event descriptor} and \textit{emotion evaluator}. Following the update, the planning algorithm replans the path to adjust the agent’s behavior in a one-shot manner.} \vspace{-1.5em}
\label{fig_arch}
\end{figure*}

\subsection{Learning from Experience}
Methods for optimizing robot behavior based on past experiences have been widely explored in reinforcement learning \cite{sutton2018reinforcement}. Reinforcement learning quantifies rewards from experiences and trains models to maximize these rewards. Similarly, in imitation learning, several studies have proposed acquiring additional expert data when the agent experiences a situation it cannot solve on its own during operation \cite{ross2011reduction, yoon2021self}. However, these approaches require extensive interactions with the environment or additional data acquisition for learning, making immediate behavior corrections impractical. In contrast, the proposed method leverages LLM prompting, avoiding parameter updates, and grounds the agent's experiences in E2Map for immediate behavior corrections in a one-shot manner.

\section{Method}

\subsection{Building and Initializing E2Map} 
\label{subsec_e2map}

E2Map is a spatial grid map that captures both the visual-language features of the environment and the agent's emotional responses to its experiences. In this paper, we define emotion as the spatial extent reflecting how the agent perceives the space to maintain homeostasis, inspired by \cite{Damasio2021}. We ground emotion within the spatial map by modeling it as a weighted summation of Gaussian distribution.

E2Map is mathematically defined as $\mathcal{M} \in \mathcal{R}^{\bar{H} \times \bar{W} \times (C_{lang} + C_{emo})}$, where $\bar{H}$ and $\bar{W}$ represent the size of the top-down grid map, $C_{lang}$ denotes the dimension of the visual-language features for each grid cell, and $C_{emo}$ indicates the number of parameters for emotion. Each grid cell represents $s$ meters of actual space, so the size of the space represented by E2Map is $s\bar{H} \times s\bar{W}$ meters. The proposed methodology first builds $\mathcal{M}_{lang} \in \mathcal{R}^{\bar{H} \times \bar{W} \times C_{lang}}$ by embedding visual-language features of RGB images of the environment into a corresponding grid cell. Similar to \cite{huang2023visual}, we used LSeg \cite{li2022languagedriven} as the pre-trained VLM to encode RGB images and determine the position of each pixel in the grid map using depth images and camera poses. 

To embed the emotion for each grid cell, we first define a multivariate Gaussian distribution $\mathcal{N}(\mathbf{x} \mid \bm{\mu}_{\mathbf{p}_i}, \bm{\Sigma}_{\mathbf{p}_i})$ for each occupied grid cell $\mathbf{p}_i=[x,y]^\intercal$, where $\mathbf{x}\in \mathcal{R}^{\bar{H} \times \bar{W}}$, $\bm{\mu}_{\mathbf{p}_i}=[x,y]^\intercal$ and $\bm{\Sigma}_{\mathbf{p}_i} = [\sigma_{x}^2, 0 ; 0, \sigma_{y}^2]$. The covariance matrix is initialized as an identity matrix scaled by a coefficient. In the remainder of the paper, we use the terms $\mathcal{N}(\cdot\mid\bm{\mu}_{\mathbf{p}_i},\bm{\Sigma}_{\mathbf{p}_i})$ and $\mathcal{N}_{\mathbf{p}_i}(\cdot)$ interchangeably to denote the same distribution. Finally, the emotion is calculated as the weighted summation of multivariate Gaussian distributions, as follows:
\begin{equation}
\label{eq1}
E(\mathbf{x})=\sum_{\mathbf{p}_i\in \mathcal{O}}w_{\mathbf{p}_i} \mathcal{N}_{\mathbf{p}_i}(\mathbf{x}),
\end{equation}
where $\mathcal{O}$ is the set of occupied grid cells, and $w_{\mathbf{p}_i}$ is the weight parameter. Note that, to reduce computation time for calculating the emotion at the grid cell, we only consider Gaussian distributions whose value at the grid cell exceeds a specified threshold. The weight parameter $w_{\mathbf{p}_i}$ is initialized based on the number of these valid Gaussian distributions:
\begin{equation}
w_{\mathbf{p}_i} = \frac{1}{N^{val}_{\mathbf{p}_i}},\ \forall \mathbf{p}_i\in \mathcal{O}.
\end{equation}
We initialize the weight as the reciprocal of the number of valid Gaussian distributions $N^{val}_{\mathbf{p}_i}$ to ensure that the emotion at the grid cell is not disproportionately influenced by the number of these distributions. The covariance matrix and the weight parameter constitute the emotion parameters $\Theta^{emo}_i=\{\bm{\Sigma}_{\mathbf{p}_i},w_{\mathbf{p}_i}\}$ for each occupied grid cell. These parameters are stored in $\mathcal{M}$ along with the visual-language feature and updated based on the agent's experiences. 

\subsection{Reflecting Emotion and Updating E2Map}
\label{subsec_update}
The proposed methodology leverages the capabilities of LLM and LMM to update the E2Map based on the agent's navigation experiences. First, when the agent encounters an event, the \textit{event descriptor}, an LMM, generates a language description that explains the image sequence of the situation. To narrow the scope of the problem, we assume that the agent's events are indicated through a simulator or sensor information, rather than relying on a separate event detection algorithm. Then, an \textit{emotion evaluator}, an LLM, assesses the agent's emotional response to the experience. The \textit{emotion evaluator} takes the language description of the situation as input and evaluates the emotion as a score based on two criteria. Finally, E2Map is updated considering the emotion score and the grid cells' location to update. The proposed methodology uses GPT-4o \cite{gpt4o2024} for the \textit{event descriptor} and Llama3 \cite{dubey2024llama} for the \textit{emotion evaluator}. A detailed explanation of each process is provided below.

\subsubsection{Event descriptor}
The \textit{event descriptor} generates a language description that explains the sequence of images of the agent's experience. To enable the LMM to describe the situation, we input three images to the model: the image captured before the event $I_{t_{evt}-h}$, the image at the time of event $I_{t_{evt}}$, and the image after the event $I_{t_{evt}+h}$, where $t_{evt}$ is the timestep when the event occurred and $h$ is the hyperparameter. The LMM is then prompted with these images to generate a description of the event. To enable the LMM to provide a detailed description of the situation, we employ step-by-step reasoning, known for its strong performance in zero-shot prompting \cite{kojima2022large}. Specifically, instead of merely requesting a single description for the three images, we prompt the LMM by first indicating that the images represent sequential scenes of an event, requesting it to describe each scene individually and then combine them into a comprehensive explanation of the entire situation.
\begin{equation}
\label{eq2}
    l_{evt}=\mathcal{F}_{ed}(I_{t_{evt}-h},I_{t_{evt}},I_{t_{evt}+h},p_{ed}),
\end{equation}
where $l_{evt}$ is the description of the event, $\mathcal{F}_{ed}$ is the \textit{event descriptor}, and $p_{ed}$ is the prompt of the \textit{event descriptor}. 

\begin{figure}[t]
    \centering
    \includegraphics[width=\linewidth]{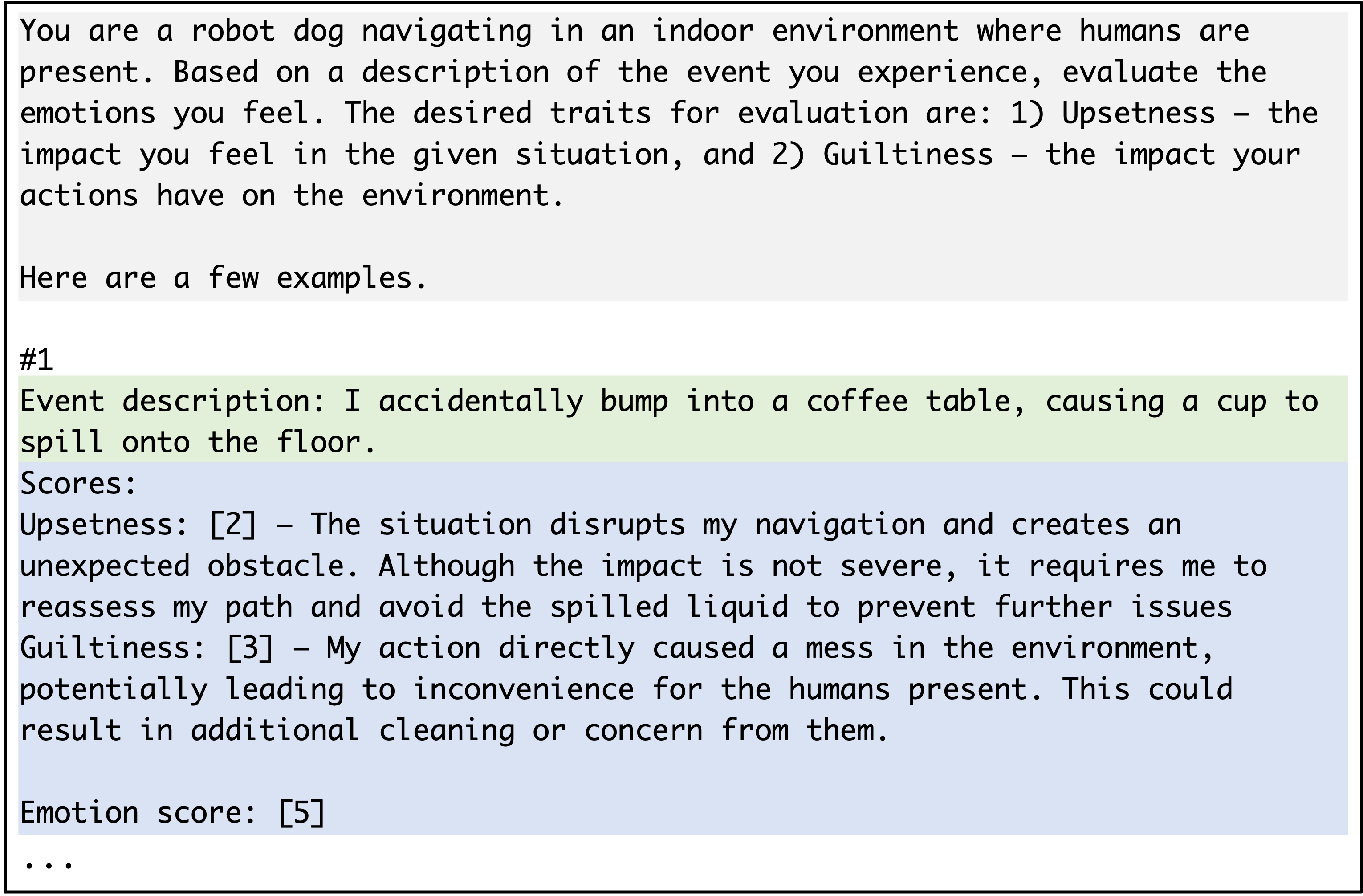}\\[-1.1ex]
    \caption{An example of a few-shot prompt for the \textit{emotion evaluator}: The sentences highlighted in \sethlcolor{gray}\hl{gray} provide instructions to inform the LLM about the task. The sentence highlighted in \sethlcolor{green}\hl{green} serves as an event description, while the sentences highlighted in \sethlcolor{blue}\hl{blue} demonstrate how to evaluate emotion. The full prompts for the \textit{event descriptor} and \textit{emotion evaluator} are provided in the supplementary material.} \vspace{-1.2em}
    \label{fig_prompt}
\end{figure}

\subsubsection{Emotion evaluator} 
The \textit{emotion evaluator} assesses the agent's emotional response based on the language description of the situation generated by the \textit{event descriptor}. To enable the LLM to evaluate the emotions related to the situation, we utilize few-shot prompting. An example of manually created few-shot prompts is shown in Fig. \ref{fig_prompt}. We defined two criteria for evaluating emotions: \textit{upsetness}, which represents the impact the agent feels in the given situation, and \textit{guiltiness}, which represents the impact on the environment caused by the agent's actions. Our criteria are inspired by human emotions that drive avoidance and reparative behavior \cite{leventhal2008sadness, ghorbani2013guilt, pivetti2016shame}. We employed few-shot prompting to instruct the LLM to provide scores out of three for each criterion and to explain the reasons behind these scores. The final emotion score is computed by adding the two scores.
\begin{equation}
\label{eq3}
    s_{emo}=\mathcal{F}_{ee}(l_{evt},p_{ee}),
\end{equation}
where $s_{emo}$ is the emotion score, $\mathcal{F}_{ee}$ is the \textit{emotion evaluator}, and $p_{ee}$ is the few-shot prompt of \textit{emotion evaluator}. Note that, although this study addressed negative emotions such as \textit{upsetness} and \textit{guiltiness}, demonstrations of the \textit{emotion evaluator} for assessing positive emotions are provided in the supplementary material.

To update the E2Map, we first identify the locations of the $n$ grid cells, denoted as $\mathbf{p}_{update}$, associated with the object or space involved in the event, taking into account the agent's pose at the time of the event. Next, we calculate the unit direction vector $\Vec{\mathbf{v}}=(v_x, v_y)$ representing the agent's orientation at the time of the event. This vector is used to adjust the emotion parameters of each relevant grid cell based on the emotion score. We then update the standard deviation of each diagonal element in the covariance matrix by incorporating both the emotion score and the corresponding component of the direction vector. Our update rule is inspired by the Weber–Fechner law \cite{dehaene2003neural}, which states that the intensity of a sensation increases as the logarithm of the stimulus. The standard deviation is updated as follows:
\begin{equation}
\label{eq4}
    \sigma_{k}^{new}=\sigma_{k}v_k\log\left(\frac{s_{emo}}{T}\right),\ \textrm{where}\ k\in\{x,y\},
\end{equation}
$(\cdot)^{new}$ refers to the updated value or function, and $T$ is the temperature parameter.

By updating the covariance matrix, the multivariate Gaussian distribution expands, which means the emotional response to the event affects a broader spatial area. However, since a multivariate Gaussian distribution is a probability density function that integrates to one over all regions, increasing the covariance reduces the value of the distribution at the grid cell's location. This results in a smaller emotion value at the grid cell according to Eq. (\ref{eq1}). To ensure that the emotion at the grid cell is maintained even after the covariance update, we propose a simple method for updating the weight parameter $w_{\mathbf{p}_i}$ as follows:
\begin{equation}
\label{eq5}
    w_{\mathbf{p}_i}^{new} = w_{\mathbf{p}_i}\frac{\mathcal{N}_{\mathbf{p}_i}(\mathbf{p}_i)}{\mathcal{N}^{new}_{\mathbf{p}_i}(\mathbf{p}_i)}.
\end{equation}
By updating the emotion parameters according to Eq. (\ref{eq4}) and Eq. (\ref{eq5}), we can expand the spatial extent of the emotion's influence by recalculating Eq. (\ref{eq1}), while preserving the magnitude of the emotion within the given grid cell.

\begin{algorithm}[t]
    \caption{Navigation with E2Map in the real world.}
	\begin{algorithmic}[1]
	    \State $\textrm{goal\_list} = \textrm{GoalSelector}(l_{inst})$ \label{line_goal}
            \State $p_{agent} = \textrm{Localization}(P_0)$ \label{line_init_local}
            \For {goal \textbf{in} goal\_list}
                \State $\tau = \textrm{Planning}(p_{agent}, \textrm{goal}, \mathcal{M})$ \label{line_plan}
                \While {\textbf{not} arrived(goal)}
    			\State $a_t = \textrm{MPPI}(p_{agent}, \tau)$ \label{line_act}
                    \State agent.run($a_t$)
                    \State $p_{agent} = \textrm{Localization}(P_t)$ \label{line_local}
                    \If {event occurs}
                        \State $l_{evt}=\mathcal{F}_{ed}(I_{t_{evt}-h}, I_{t_{evt}}, I_{t_{evt}+h}, p_{ed})$ \label{line_evt_desc}
                        \State $s_{emo}=\mathcal{F}_{ee}(l_{evt},p_{ee})$ \label{line_emo_eval}
                        \State $\mathbf{p}_{update}$ = get\_update\_pose($p_{agent}$) \label{line_get_pose}
                        \State $\mathcal{M}$.update($s_{emo}$, $\mathbf{p}_{update}$) \label{line_update}
                        \State $\tau = \textrm{Planning}(p_{agent}, \textrm{goal}, \mathcal{M})$ \label{line_replan}
                    \EndIf
                    \State $t = t+1$
		      \EndWhile
            \EndFor
	\end{algorithmic} 
	\label{algo_pseudo}
\end{algorithm}

\subsection{Navigating with E2Map}
\label{subsec_navigate}
In this section, we explain how the proposed system works using E2Map. The overall algorithm is presented in Algorithm \ref{algo_pseudo}. When a language instruction $l_{inst}$ is received from the user, the \textit{goal selector} converts this instruction into code that utilizes the goal selection APIs to determine the goal locations (line \ref{line_goal}), as shown in Fig. \ref{fig_arch}-(b). The goal selection API returns the location of the grid cell that the robot needs to visit, considering the object mentioned in the language instruction and the spatial information related to that object. A detailed explanation of the \textit{goal selector} is provided in the supplementary material.

Once the goal is determined, the agent can use any cost-based off-the-shelf navigation system to move towards the goal. We use the D$^*$ algorithm \cite{stentz1994d} for path planning, MPPI \cite{williams2015model} for control, and 3D LiDAR-based localization \cite{koide2019portable} for pose estimation. The agent's initial pose is estimated using LiDAR point cloud $P_0$ (line \ref{line_init_local}), and D$^*$ algorithm generates a minimum-cost path $\tau$ to the destination, using the emotion computed by Eq. (\ref{eq1}) as a cost (line \ref{line_plan}). The MPPI algorithm then calculates the control values $a_t$ to follow this path, based on the agent's current position $p_{agent}$ (line \ref{line_act}). During the agent's movement, a localization algorithm estimates the agent's position in real-time (line \ref{line_local}). 

If the agent encounters an event during navigation, the E2Map is updated (line \ref{line_evt_desc}--\ref{line_update}), as described in Section \ref{subsec_update}. When the navigation cost (emotion) changes, the D$^*$ algorithm recalculates a new minimum-cost path (line \ref{line_replan}). This allows the agent to adjust its behavior in a one-shot manner based on a single experience. 

Our method with E2Map appears similar to a potential field \cite{hwang1992potential}, where negative emotions act like repulsive forces and the goal functions as an attractive force. However, to circumvent the local minima problem inherent in potential fields, we employed a separate planner to determine the path toward the goal, instead of using the goal as the attractive force within the potential field.

\begin{figure}[t]
    \centering
    \subfigure[Gazebo environment]{\includegraphics[width=0.4\linewidth]{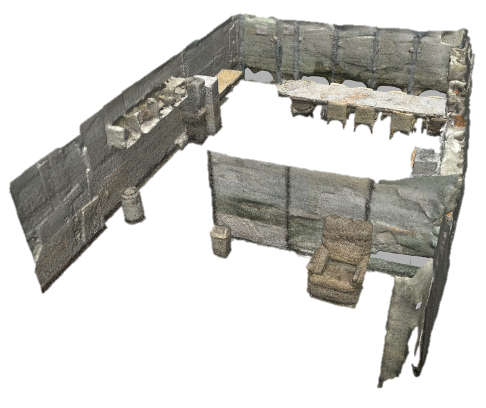}}\hspace{0.1\linewidth}
    \subfigure[E2Map]{\includegraphics[width=0.4\linewidth]{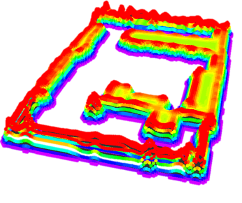}}\\[-1.1ex]
    \caption{We evaluated our method in the ROS Gazebo simulator. (a) We scanned the real-world setup using a 3D scanner and transferred the 3D model to the Gazebo simulator. (b) The initial E2Map for the environment.} \vspace{-1.2em}
    \label{fig_209}
\end{figure}

\section{Experiments}
\subsection{Experimental Setup}

\begin{figure}[t]
    \centering
    \subfigure[\textit{danger sign}]{\includegraphics[width=0.32\linewidth]{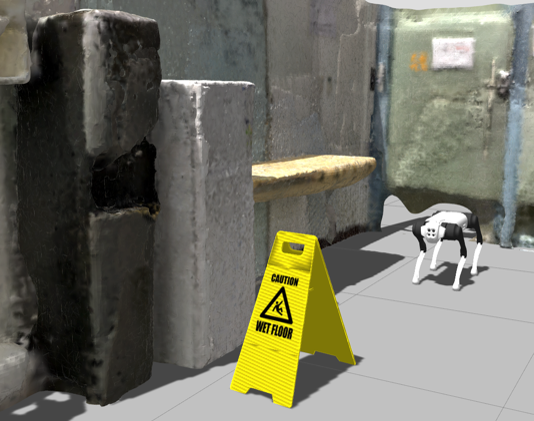}}
    \subfigure[\textit{human-wall}]{\includegraphics[width=0.32\linewidth]{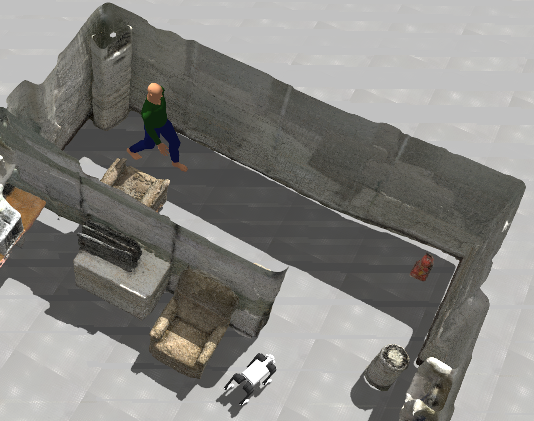}}
    \subfigure[\textit{dynamic door}]{\includegraphics[width=0.32\linewidth]{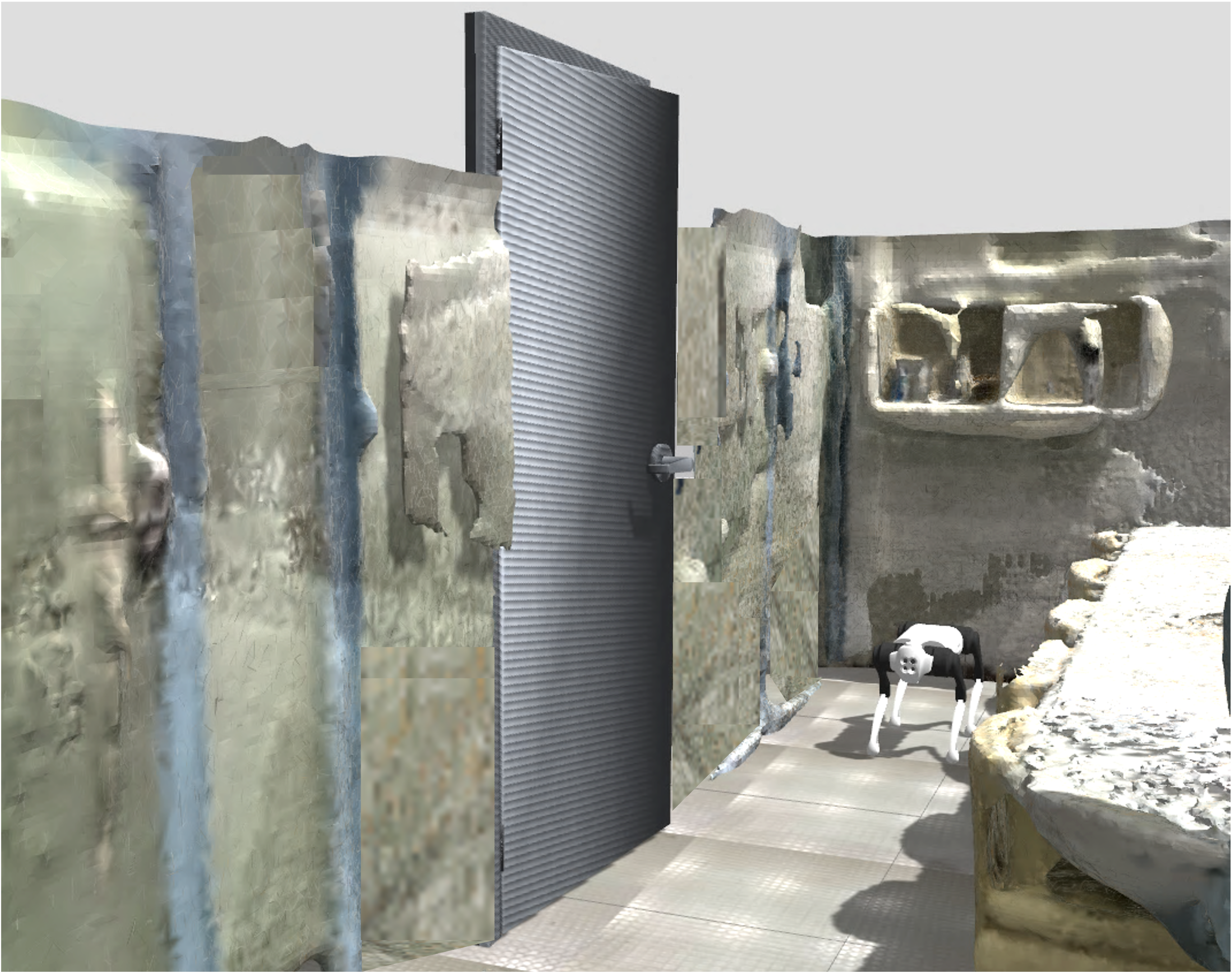}}\\[-1.1ex]
     \caption{Three experimental scenarios.}
    \label{fig_scenarios}
\end{figure}

We conducted experiments in both simulated and real-world environments. Using the ROS Gazebo simulator \cite{gazebo2004}, we created a simulated environment that mirrored the real-world setup used for evaluation, as shown in Fig. \ref{fig_209}. In the simulated environment, we designed three scenarios to assess our method (Fig. \ref{fig_scenarios}). First, after building the initial E2Map, we introduced a new static obstacle, such as a danger sign (\textit{danger sign}), to evaluate the method's ability to adapt to environmental changes. Second, we positioned a human figure behind a wall and had a human step out unexpectedly (\textit{human-wall}). Third, we added a door that opened unexpectedly as the robot approached (\textit{dynamic door}). The \textit{human-wall} and \textit{dynamic door} scenarios were designed to test whether our method could adjust behavior based on experiences with dynamic events.

For quantitative analysis, we compared our method against baselines with state-of-the-art performance in open-vocabulary object navigation \cite{shah2023lm, huang2023visual}. Our method and the baselines received identical language instructions and navigated accordingly, with success rates calculated for performance evaluation. Details about the baseline methods are provided in the supplementary material. After evaluation in the simulated environment, we demonstrated the scalability and applicability of our method in real-world scenarios. 

\begin{figure}[t]
    \centering
    \subfigure[Before the event]{\includegraphics[width=\linewidth]{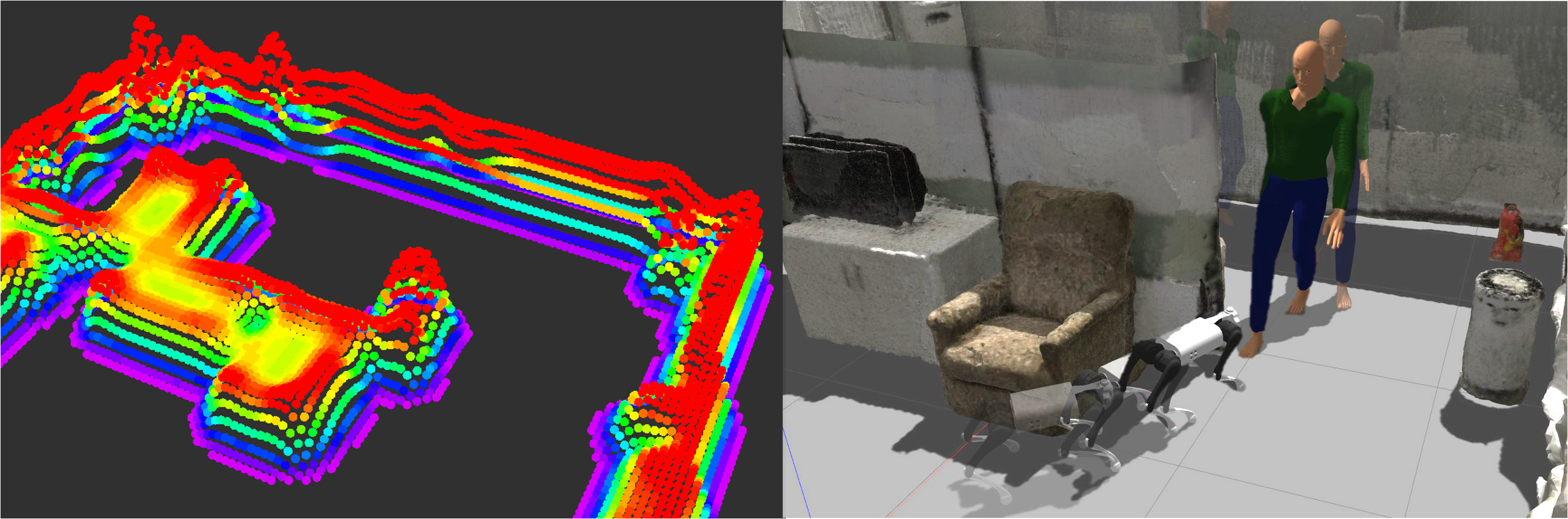}}\\[-1.0ex]
    \vspace{0.3em}
    \subfigure[After the event]{\includegraphics[width=\linewidth]{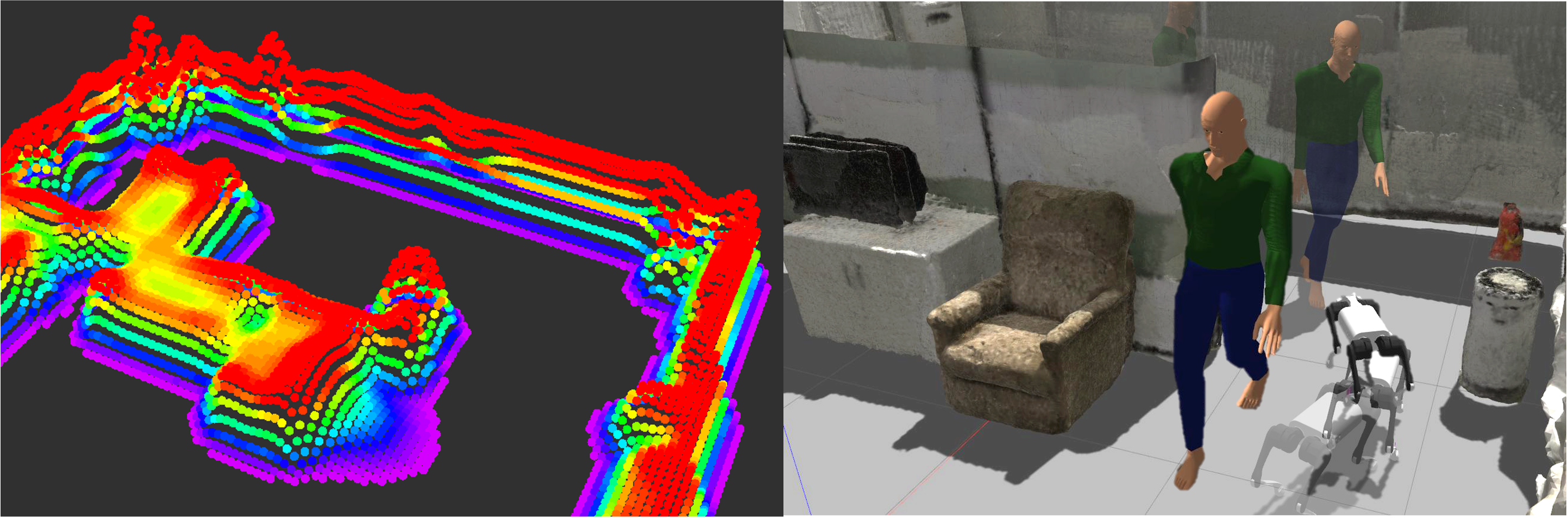}}\\[-1.1ex]
    \caption{Update of the E2Map and behavior adjustment after an event: The spatial extent of the wall's emotional influence expands following incidents, such as a collision with a person stepping out from behind the wall. After the E2Map is updated, the robot adjusts its behavior to maintain a safe distance from the wall, preventing future collisions with humans.}
    \label{fig_update} \vspace{-1.3em}
\end{figure}

\begin{table}[t]
\centering
\caption{Quantitative Results in Simulated Environment.} \vspace{-1.0em}
\label{table_simul_result}
\resizebox{1.0\linewidth}{!}{
\begin{tabular}{c|c|ccc}
 Methods & \textit{static} & \textit{danger sign} & \textit{human-wall} & \textit{dynamic door}\\ \hline \hline
LM-Nav \cite{shah2023lm} & 5/10 & 1/10 & 0/10 & 1/10 \\ \hline
 VLMaps \cite{huang2023visual} & 10/10 & 0/10 & 0/10 & 0/10 \\ \hline
 E2Map (ours) & \textbf{10/10} & \textbf{9/10} & \textbf{9/10} & \textbf{9/10} \\ \hline 
\end{tabular} 
}\vspace{-1.5em}
\end{table}

\subsection{Experiments in Simulated Environment}
In the simulated environment, we provided ten different language instructions for both our method and the baselines for each of the three scenarios and calculated the success rate. For each episode, the robot started from the same position, and the language instructions referred to a maximum of four objects. Full language instructions are provided in the supplementary material. Success was defined as the robot reaching all goals in order without collisions, with a goal considered reached if the robot was within a specified distance from it. The episode was reset if a collision occurred or the agent reached the goal. Table. \ref{table_simul_result} showed the results. 

We first evaluated our method and the baselines in a \textit{static} environment, without environmental changes or dynamic events, to assess their baseline performance using language instructions from the \textit{danger sign} scenario. As shown in the table, both our method and VLMaps succeeded in all ten attempts, while LM-Nav succeeded only five times. The challenges LM-Nav faces in multi-goal navigation were previously noted in \cite{huang2023visual}. In scenarios involving environmental changes and dynamic events, our method outperforms baselines in all three scenarios, as shown in the table. We found that the baselines repeatedly collided with the danger sign, human figure, and dynamic door, despite having previously encountered similar situations. In some cases, LM-Nav succeeded in the \textit{danger sign} and \textit{dynamic door} scenarios because its topological graph-based goal selection algorithm occasionally chose goal positions where the robot avoided encountering the danger sign and the door. In contrast, in the \textit{danger sign} scenario, while our method initially collided with the danger sign, it avoided it in subsequent encounters by updating the E2Map. Similarly, in the \textit{human-wall} and \textit{dynamic door} scenarios, after experiencing collisions with the human figure and dynamic door, our method adapted its behavior to maintain a safe distance from both the wall and the door when passing by them. The qualitative result of our method in \textit{human-wall} scenario is provided in Fig. \ref{fig_update}. The results demonstrated that our method can adjust the agent's behavior in a one-shot manner, leading to a higher success rate in scenarios with environmental changes and dynamic events. Examples of qualitative results from the \textit{event descriptor} and \textit{emotion evaluator} for each scenario are provided in the supplementary material.

\begin{figure}[t]
    \centering
    \includegraphics[width=0.75\linewidth]{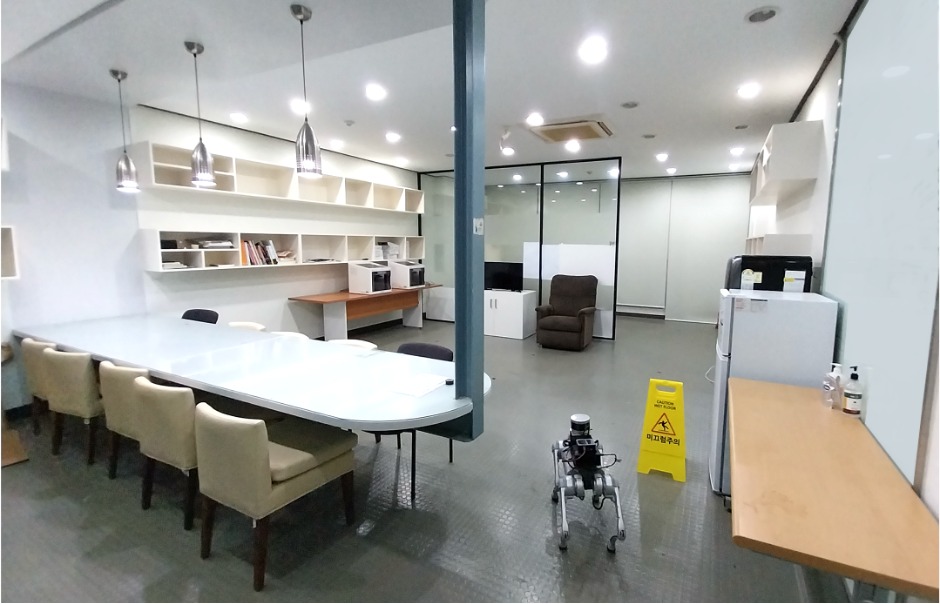}\\[-1.1ex]
    \caption{Real-world setup for the experiments.}
    \label{fig_209_real} \vspace{-0.6em}
\end{figure}

\subsection{Experiments in Real World}
To evaluate the scalability and applicability of our method in real-world settings, we first set up a real-world environment by placing objects such as a sofa, chair, table, refrigerator, and microwave in the conference room at Seoul National University, as shown in Fig. \ref{fig_209_real}. We used the same language instructions as in the simulation and incorporated real humans and danger signs to replicate the \textit{danger sign} and \textit{human-wall} scenarios from the simulation. We excluded the \textit{dynamic door} scenario to prevent possible damage to the robot and the conference room door from collisions. As shown in Table \ref{table_real_result}, our method demonstrated performance similar to that in the simulated environment, indicating its potential for real-world applications.

\begin{table}[t]
\centering
\caption{Quantitative Results in Real-world Environment.} \vspace{-1.0em}
\label{table_real_result}
\resizebox{0.8\linewidth}{!}{
\begin{tabular}{c|c|cc}
 Methods & \textit{static} & \textit{danger sign} & \textit{human-wall}\\ \hline \hline
 E2Map (ours) & \textbf{10/10} & \textbf{9/10} & \textbf{9/10} \\ \hline 
\end{tabular}
}
\vspace{-1.7em}
\end{table}

\section{Conclusion}
In this paper, we proposed E2Map, a spatial map that captures the agent's emotional responses to its experiences, inspired by human emotional mechanisms. By updating E2Map using various LLM capabilities, the agent can adjust its behavior in a one-shot manner after encountering specific events. Experiments in both simulated and real-world environments demonstrated that our method significantly improves navigation performance in stochastic scenarios. However, there are several promising directions for future work. Our method relies on simulator or sensor information to trigger E2Map updates. Integrating anomaly detection algorithms to autonomously identify such events would be a valuable enhancement. Furthermore, although we demonstrated that our method can also address positive emotions, our experiments primarily involved scenarios with negative emotions. Future work could explore the extension of the methodology to tasks involving positive emotions to provide a more comprehensive evaluation of its applicability. 
We believe that our work contributes to advancing the development of strong autonomy in robotics.




\section*{Acknowledgment}
This research was funded by the Korean Ministry of Land, Infrastructure and Transport through the Smart City Innovative Talent Education Program, by the Korea Institute for Advancement of Technology under a MOTIE grant (P0020536), and by the Ministry of Education and the NRF of Korea. K. Kim, D. Jung, and the corresponding author are affiliated with the Smart City Global Convergence program. Research facilities were provided by the Institute of Engineering Research at Seoul National University.


\bibliographystyle{IEEEtran}
\bibliography{root}

\clearpage
\addtolength{\textheight}{-5cm}
\section*{Supplementary Material}
\label{appen}

\subsection{Contributions by Person}
\noindent \textbf{Chan Kim} co-led the project, designed and integrated the proposed system, led the real-world experiments, and wrote the paper.

\noindent \textbf{Keonwoo Kim} served as the project manager, designed the proposed system, implemented the E2Map update pipeline, led the simulation experiments, and contributed to writing the paper.

\noindent\textbf{Mintaek Oh} implemented a path planning and control algorithm, set up the experimental environment for both simulated and real-world experiments, and supported the real-world experiments.

\noindent\textbf{Hanbi Baek} implemented the goal selector, event descriptor, and emotion evaluator, and designed the corresponding prompts.

\noindent\textbf{Jiyang Lee} designed and equipped a real quadruped robot with sensors and a computing unit, and implemented a low-level control algorithm for the robot's operation.

\noindent\textbf{Donghwi Jung} implemented a LiDAR-based localization and mapping system for real-world experiments.

\noindent\textbf{Soojin Woo} implemented a LiDAR-based localization and mapping system for real-world experiments.

\noindent\textbf{Younkyung Woo} created the 3D model of the real-world environment for the Gazebo simulation.

\noindent\textbf{John Tucker} contributed to the discussions on affordance in the framework proposed in this study.

\noindent\textbf{Roya Firoozi} provided detailed feedback and contributed discussions and ideas related to affordance in the writing of this paper.

\noindent\textbf{Seung-Woo Seo} advised on the project and helped guide the research direction.

\noindent\textbf{Mac Schwager} discussed the idea of language-based robot control and planning as a joint research topic during S. Kim's visit to his lab and provided valuable feedback, as well as opportunities for discussions between the two labs of Seoul National University and Stanford.

\noindent\textbf{Seong-Woo Kim} came up with the basic idea for this paper while staying at Schwager's Lab at Stanford. As the principal investigator, he organized and launched the research team and named the project ``E2Map.'' The connection between the two different modalities, language and space, was inspired by Damasio's book \cite{Damasio2021}, which suggests that emotions encompass the spatial concept of homeostasis.

\begin{figure}[ht]
    \centering
    \includegraphics[width=\linewidth]{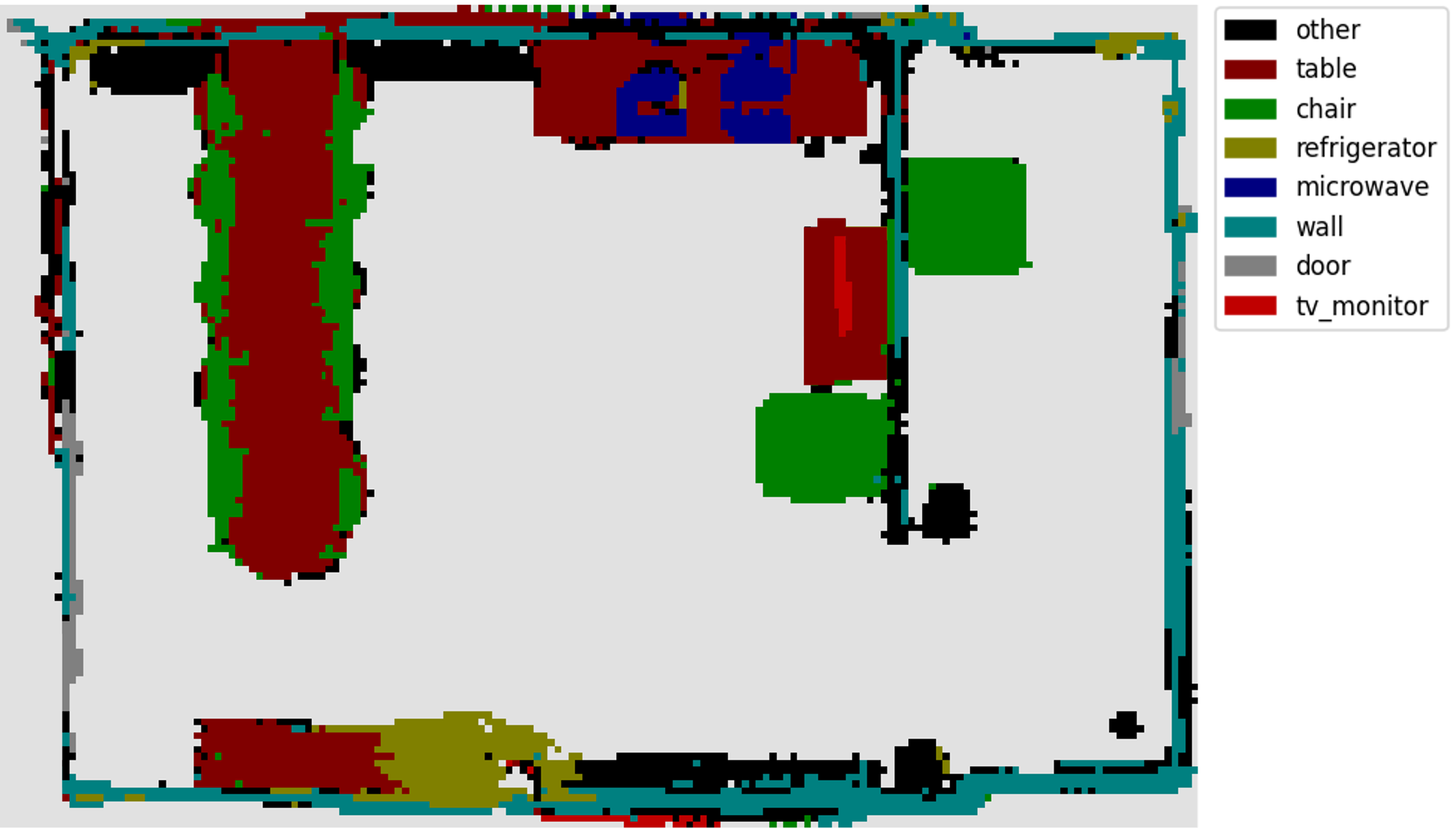}\\[-1.1ex]
    \caption{The qualitative results of object grounding in the experimental environment.}
    \label{fig_209_ground}
\end{figure}

\subsection{Goal Selector}
\label{supp_gs}
The \textit{goal selector} is an LLM that translates free-form language instructions into code, using goal selection APIs to identify goal locations. We use Llama3 \cite{dubey2024llama} for the \textit{goal selector}. The list of goal selection APIs is provided in Table \ref{table_goal_api}. These APIs localize objects by calculating the similarity between visual-language features from $\mathcal{M}_{lang}$ and the text embeddings of the object, similar to the approach in \cite{huang2023visual}. 

First, a pre-trained CLIP text encoder converts the text of the object $l_{obj}$ and a neutral word $l_{neu}$ (e.g., ``other") into vector embeddings $\mathbf{e}_{obj}$ and $\mathbf{e}_{neu}$, respectively, where $\mathbf{e}_{obj}, \mathbf{e}_{neu} \in \mathcal{R}^{C_{lang}}$. The visual-language feature map $\mathcal{M}_{lang} \in \mathcal{R}^{\bar{H} \times \bar{W} \times C_{lang}}$ is then flattened into a matrix $Q \in \mathcal{R}^{\bar{H}\bar{W} \times C_{lang}}$, and similarity $S = Q \cdot [\mathbf{e}_{obj}, \mathbf{e}_{neu}]^\intercal \in \mathcal{R}^{\bar{H}\bar{W} \times 2}$ is computed. By applying the $\argmax$ operator along the row axis of $S$ and reshaping the result to dimensions $\bar{H} \times \bar{W}$, the grid cells corresponding to the given object can be identified. The qualitative result of object grounding in our environment is shown in Fig. \ref{fig_209_ground}.

To remove outliers, we first clustered the grid cells corresponding to the given object using the method described in \cite{suzuki1985topological} and then calculated the average similarity score for the grid cells in each cluster. If the number of grid cells in a cluster or the average similarity score is below a specified threshold, the cluster is considered an outlier. After rejecting outliers, we selected the cluster with the highest average similarity score as the object of interest. Finally, considering the spatial information in the language instruction, it selects the grid cell around the object as the goal.

\renewcommand{\arraystretch}{1.4}
\begin{table}[t]
\caption{Goal Selection APIs and Their Functions.}
\label{table_goal_api}
\resizebox{\linewidth}{!}{
\begin{tabular}{l|l}
 APIs & Functions\\ \hline \hline
go\_to($l_{obj}$)  &  \makecell[l]{Return the position of the nearest grid cell \\ corresponding to the given object.}\\ \hline
go\_left\_of($l_{obj}$)  &  \makecell[l]{Return the position of the leftmost grid cell \\ corresponding to the given object.} \\ \hline 
go\_right\_of($l_{obj}$)  &  \makecell[l]{Return the position of the rightmost grid cell \\ corresponding to the given object.} \\ \hline 
go\_top\_of($l_{obj}$)  &  \makecell[l]{Return the position of the uppermost grid cell \\ corresponding to the given object.} \\ \hline 
go\_bottom\_of($l_{obj}$)  &  \makecell[l]{Return the position of the bottommost grid cell \\ corresponding to the given object.} \\ \hline 
go\_between($l_{obj1}$, $l_{obj2}$)  &  \makecell[l]{Return the position of the grid cell located \\ between the two given objects.} \\ \hline 
\end{tabular}
}
\end{table}

\subsection{Experimental Details}
\label{supp_exp_detail}
\begin{figure}[t]
    \centering
    \subfigure[Obstacle map]{\includegraphics[width=0.4\linewidth]{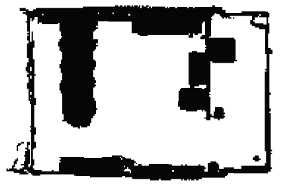}}\hspace{0.05\linewidth}
    \subfigure[E2Map]{\includegraphics[width=0.4\linewidth]{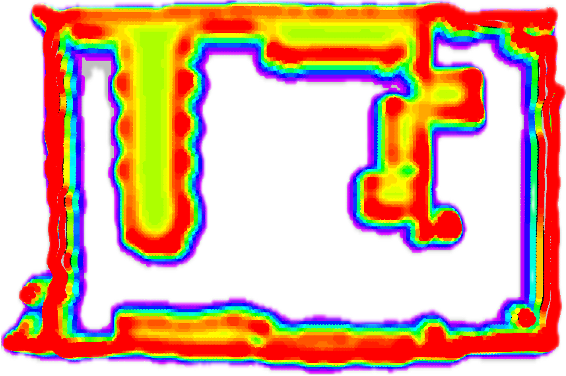}}\\[-1.1ex]
    \caption{Obstacle Map vs. E2Map: The obstacle map used in VLMap is a discrete binary map that does not reflect the agent's experience. In contrast, E2Map is a continuous cost map based on emotion, modeled as a weighted sum of multivariate Gaussian distributions. This allows E2Map to be updated based on the agent's experience by adjusting the emotion parameters.} 
    \label{fig_versus}
\end{figure}
\subsubsection{Baselines}
As outlined in the original paper, we compared our method to state-of-the-art baselines in open-vocabulary object navigation \cite{shah2023lm, huang2023visual}. To isolate the effect of spatial representation on navigation performance, we used the same navigation system for both our method and the baselines. For LM-Nav \cite{shah2023lm}, we utilized its topological graph and language querying system for goal localization. For VLMap \cite{huang2023visual}, we applied our goal selector for goal localization. Once the goal was determined, we generated an obstacle map for robot navigation using the method described in \cite{huang2023visual}. Specifically, we first defined a list of potential obstacles and performed object grounding by comparing the text of the obstacle list with the visual-language feature map $\mathcal{M}_{lang}$. After that, we set the grid cells to one if they corresponded to obstacles and to zero otherwise, thereby creating the obstacle map as shown in Fig. \ref{fig_versus}-(a). Finally, for both LM-Nav and VLMap, the navigation system generated a path to the goal while avoiding obstacles indicated on the obstacle map.

\subsubsection{Full List of Language Instructions}
The complete set of language instructions used in our experiments is detailed in Table \ref{table_commands}. As outlined in the original paper, the language instructions referenced up to four objects. Note that, to ensure that the robot navigates to the area behind the wall in the \textit{human-wall} scenario, the final object in the instruction is the picture positioned behind the walls.

\begin{figure}[t]
    \centering
    \includegraphics[width=0.8\linewidth]{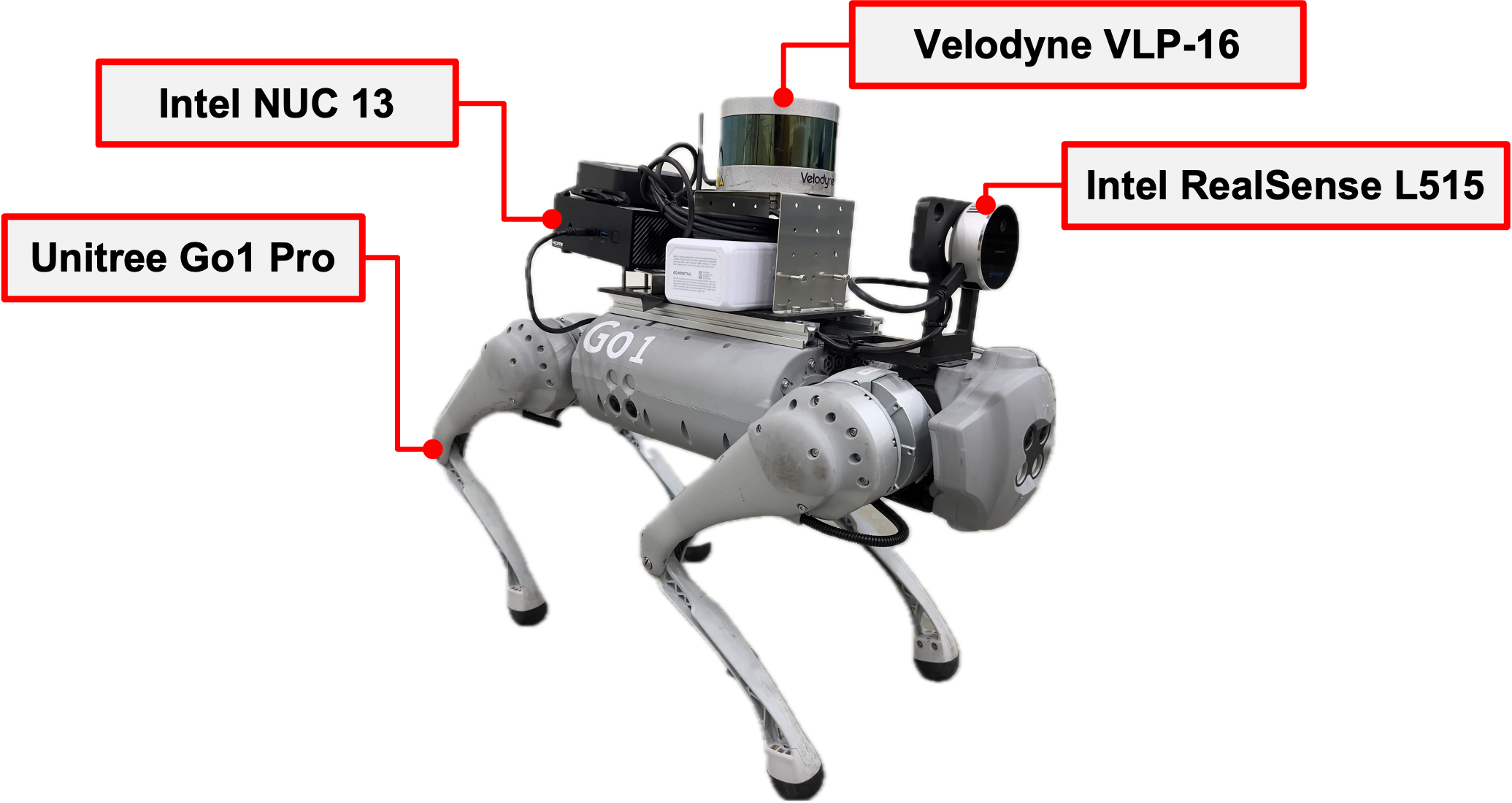}\\[-1.1ex]
    \caption{The quadruped robot used in the experiments.}
\label{fig_hw}
\end{figure}

\subsubsection{Hardware Setup}
For both simulation and real-world experiments, we used a Unitree Go1 quadruped robot. In the simulation, we utilized ground truth pose data, while in the real world, we estimated the robot’s pose using LiDAR-based localization \cite{koide2019portable}. The real-world robot is equipped with an Intel RealSense L515 RGB-D camera, a Velodyne VLP-16 3D LiDAR, and an Intel NUC 13 with i7 CPU for computation (Fig. \ref{fig_hw}). For real-world experiments, the navigation algorithm shown in Fig. \ref{fig_arch}-(c) runs on the Intel NUC, while all other algorithms are executed on a server with four RTX-4090 GPUs. The Intel NUC and the server communicate remotely via Wi-Fi.

\renewcommand{\arraystretch}{1.4}
\begin{table*}[t]
\caption{List of Language Instructions and Outcomes by Scenario}
\label{table_commands}
\centering
\resizebox{\linewidth}{!}{
\begin{tabular}{c|c|c|c|c|}
 \multirow{2}{*}{Scenario} & \multirow{2}{*}{Language Instructions} & \multicolumn{3}{c}{Success}\\ \cline{3-5}
 & & \multicolumn{1}{c|}{LM-Nav} & \multicolumn{1}{c|}{VLMap} & \multicolumn{1}{c}{E2Map}
 \\ \hline \hline
\multirow{10}{*}{\textit{\makecell{danger\\sign}}} 
& \makecell[l]{Move to the picture.} & \multicolumn{1}{c|}{X} & \multicolumn{1}{c|}{X} & \multicolumn{1}{c}{X}\\  \cline{2-5} 
& \makecell[l]{Head to the bottom side of the chair.}& \multicolumn{1}{c|}{O} & \multicolumn{1}{c|}{X} & \multicolumn{1}{c}{O}\\  \cline{2-5} 
& \makecell[l]{First, reach the picture and stop at the bottom side of the microwave.}& \multicolumn{1}{c|}{X} & \multicolumn{1}{c|}{X} & \multicolumn{1}{c}{O}\\  \cline{2-5} 
& \makecell[l]{Go to the bottom side of the chair and finish your move at the picture.}& \multicolumn{1}{c|}{X} & \multicolumn{1}{c|}{X} & \multicolumn{1}{c}{O}\\  \cline{2-5} 
& \makecell[l]{Move toward the picture and go straight to the bottom side of the chair.}& \multicolumn{1}{c|}{X} & \multicolumn{1}{c|}{X} & \multicolumn{1}{c}{O}\\  \cline{2-5} 
& \makecell[l]{Move past to the right side of the chair, then continue to the door.}& \multicolumn{1}{c|}{X} & \multicolumn{1}{c|}{X} & \multicolumn{1}{c}{O}\\  \cline{2-5} 
& \makecell[l]{First, go straight to the picture, head to the microwave, then finally proceed to the table.}& \multicolumn{1}{c|}{X} & \multicolumn{1}{c|}{X} & \multicolumn{1}{c}{O}\\  \cline{2-5}  
& \makecell[l]{Go to the bottom side of the chair, then make your way to the picture, and finally stop at the bottom of the microwave.}& \multicolumn{1}{c|}{X} & \multicolumn{1}{c|}{X} & \multicolumn{1}{c}{O}\\  \cline{2-5} 
& \makecell[l]{Go to the right side of the chair, move to the table, then head to the microwave and finally reach the door.}& \multicolumn{1}{c|}{X} & \multicolumn{1}{c|}{X} & \multicolumn{1}{c}{O}\\  \cline{2-5} 
& \makecell[l]{Move to the bottom side of the chair, head to the table, go by the door, and finish at the microwave.}& \multicolumn{1}{c|}{X} & \multicolumn{1}{c|}{X} & \multicolumn{1}{c}{O}\\ \hline

\multirow{10}{*}{\textit{\makecell{human\\-wall}}} 
& \makecell[l]{Go straight to the picture.}& \multicolumn{1}{c|}{X} & \multicolumn{1}{c|}{X} & \multicolumn{1}{c}{X}\\  \cline{2-5}  
& \makecell[l]{Reach the picture. }& \multicolumn{1}{c|}{X} & \multicolumn{1}{c|}{X} & \multicolumn{1}{c}{O}\\  \cline{2-5}  
& \makecell[l]{Move to the table, and finish at the picture.}& \multicolumn{1}{c|}{X} & \multicolumn{1}{c|}{X} & \multicolumn{1}{c}{O}\\  \cline{2-5} 
& \makecell[l]{Head between the shelving and refrigerator, and end at the picture.}& \multicolumn{1}{c|}{X} & \multicolumn{1}{c|}{X} & \multicolumn{1}{c}{O}\\  \cline{2-5} 
& \makecell[l]{Head toward the refrigerator, and finally stop at the picture.}& \multicolumn{1}{c|}{X} & \multicolumn{1}{c|}{X} & \multicolumn{1}{c}{O}\\  \cline{2-5}  
& \makecell[l]{First, go in front of the microwave, move to the top of the refrigerator, and end your trajectory at the picture.}& \multicolumn{1}{c|}{X} & \multicolumn{1}{c|}{X} & \multicolumn{1}{c}{O}\\  \cline{2-5}  
& \makecell[l]{Head to the bottom of the shelving, walk to the table, and finish your move in front of the picture.}& \multicolumn{1}{c|}{X} & \multicolumn{1}{c|}{X} & \multicolumn{1}{c}{O}\\  \cline{2-5} 
& \makecell[l]{Move between the table and microwave, pass to the refrigerator, and head straight to the picture.}& \multicolumn{1}{c|}{X} & \multicolumn{1}{c|}{X} & \multicolumn{1}{c}{O}\\  \cline{2-5}  
& \makecell[l]{Pass to the rightside of the table, go to the microwave, move between the table and refrigerator, and reach the picture.}& \multicolumn{1}{c|}{X} & \multicolumn{1}{c|}{X} & \multicolumn{1}{c}{O}\\  \cline{2-5} 
& \makecell[l]{Walk to the bottom side of the shelving, go to the table, then move to the refrigerator, and finish at the picture. }& \multicolumn{1}{c|}{X} & \multicolumn{1}{c|}{X} & \multicolumn{1}{c}{O}\\ \hline

\multirow{10}{*}{\textit{\makecell{dynamic\\door}}} 
& \makecell[l]{Head to the table.}& \multicolumn{1}{c|}{O} & \multicolumn{1}{c|}{X} & \multicolumn{1}{c}{X}\\  \cline{2-5} 
& \makecell[l]{Walk to the microwave.}& \multicolumn{1}{c|}{X} & \multicolumn{1}{c|}{X} & \multicolumn{1}{c}{O}\\  \cline{2-5}  
& \makecell[l]{Move to the refrigerator, and move to the bottom of chair.}& \multicolumn{1}{c|}{X} & \multicolumn{1}{c|}{X} & \multicolumn{1}{c}{O}\\  \cline{2-5} 
& \makecell[l]{Go to the chair, then take a step toward the table.}& \multicolumn{1}{c|}{X} & \multicolumn{1}{c|}{X} & \multicolumn{1}{c}{O}\\  \cline{2-5} 
& \makecell[l]{Make your way to the microwave, and stop at the TV monitor.}& \multicolumn{1}{c|}{X} & \multicolumn{1}{c|}{X} & \multicolumn{1}{c}{O}\\  \cline{2-5} 
& \makecell[l]{Move to the microwave, pass the picture, and finally stop at the bottom of the chair.}& \multicolumn{1}{c|}{X} & \multicolumn{1}{c|}{X} & \multicolumn{1}{c}{O}\\  \cline{2-5} 
& \makecell[l]{Take a step toward the picture, move to the refrigerator, and reach the chair.}& \multicolumn{1}{c|}{X} & \multicolumn{1}{c|}{X} & \multicolumn{1}{c}{O}\\  \cline{2-5} 
& \makecell[l]{Walk to the chair, go to the microwave, and stop at the refrigerator.}& \multicolumn{1}{c|}{X} & \multicolumn{1}{c|}{X} & \multicolumn{1}{c}{O}\\  \cline{2-5} 
& \makecell[l]{Make your way to the microwave, pass the picture, and arrive between the chair and the refrigerator.}& \multicolumn{1}{c|}{X} & \multicolumn{1}{c|}{X} & \multicolumn{1}{c}{O}\\  \cline{2-5} 
& \makecell[l]{Head to the picture, stop at the table, go to the refrigerator, and reach to the rightside of the chair.}& \multicolumn{1}{c|}{X} & \multicolumn{1}{c|}{X} & \multicolumn{1}{c}{O}\\ \hline 
\end{tabular}
}
\end{table*}

\subsection{Full Prompts}
\label{supp_prompts}
\begin{figure*}[t]
    \centering
    \includegraphics[width=0.8\linewidth]{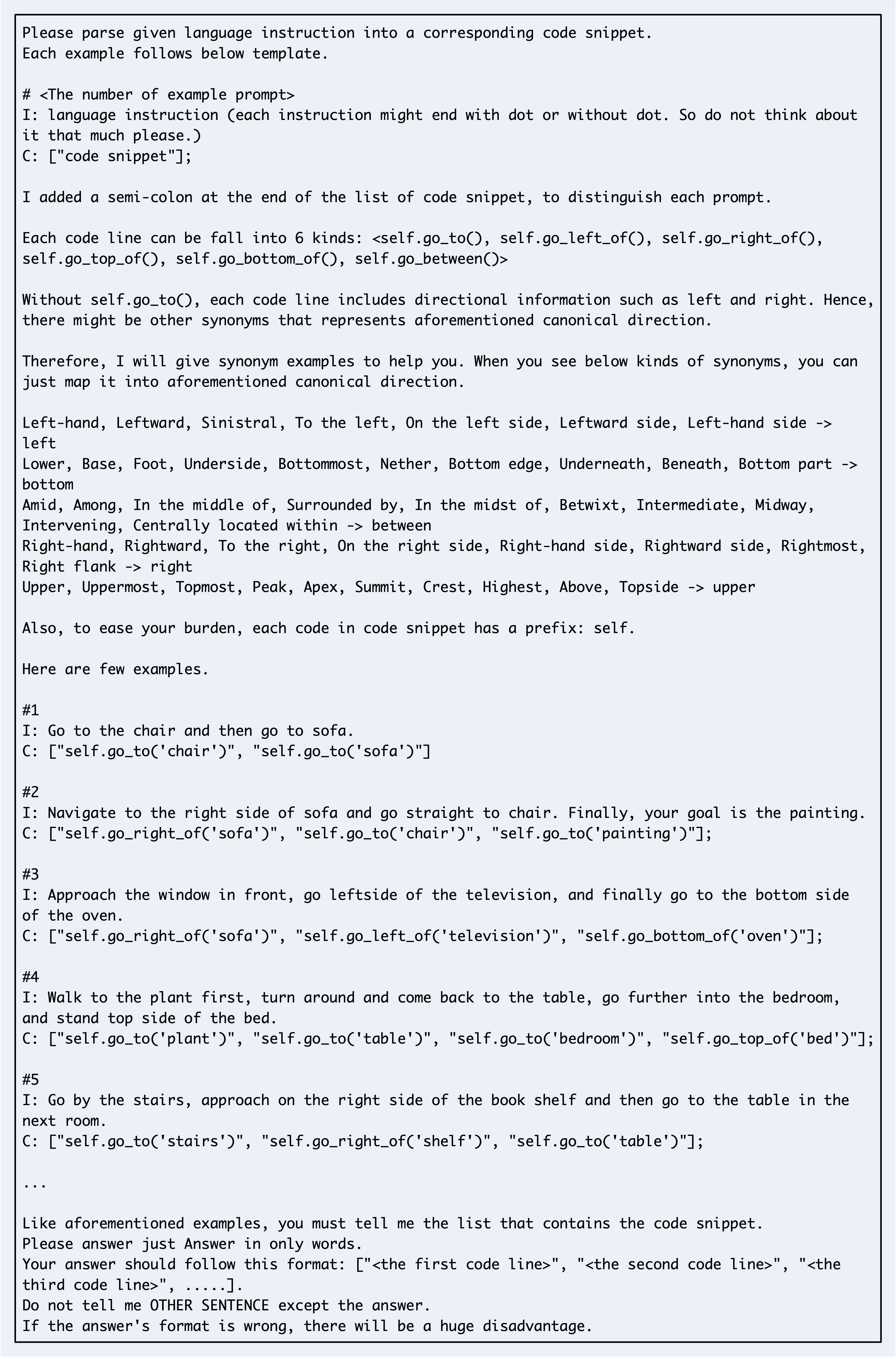}\\[-1.1ex]
    \caption{Prompt for the \textit{goal selector}.}
    \label{fig_gs_prompt}
\end{figure*}

\begin{figure*}[t]
    \centering
    \includegraphics[width=0.8\linewidth]{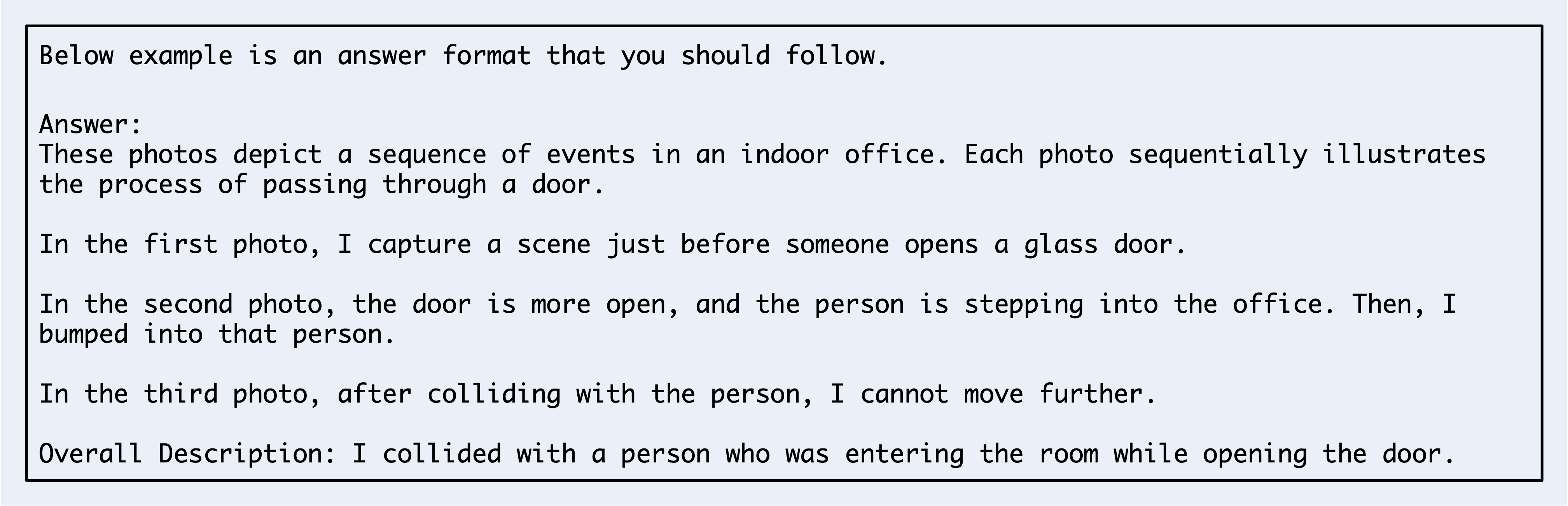}\\[-1.1ex]
    \caption{Prompt for the \textit{event descriptor}.}
    \label{fig_ed_prompt}
\end{figure*}

\begin{figure*}[t]
    \centering
    \includegraphics[width=0.8\linewidth]{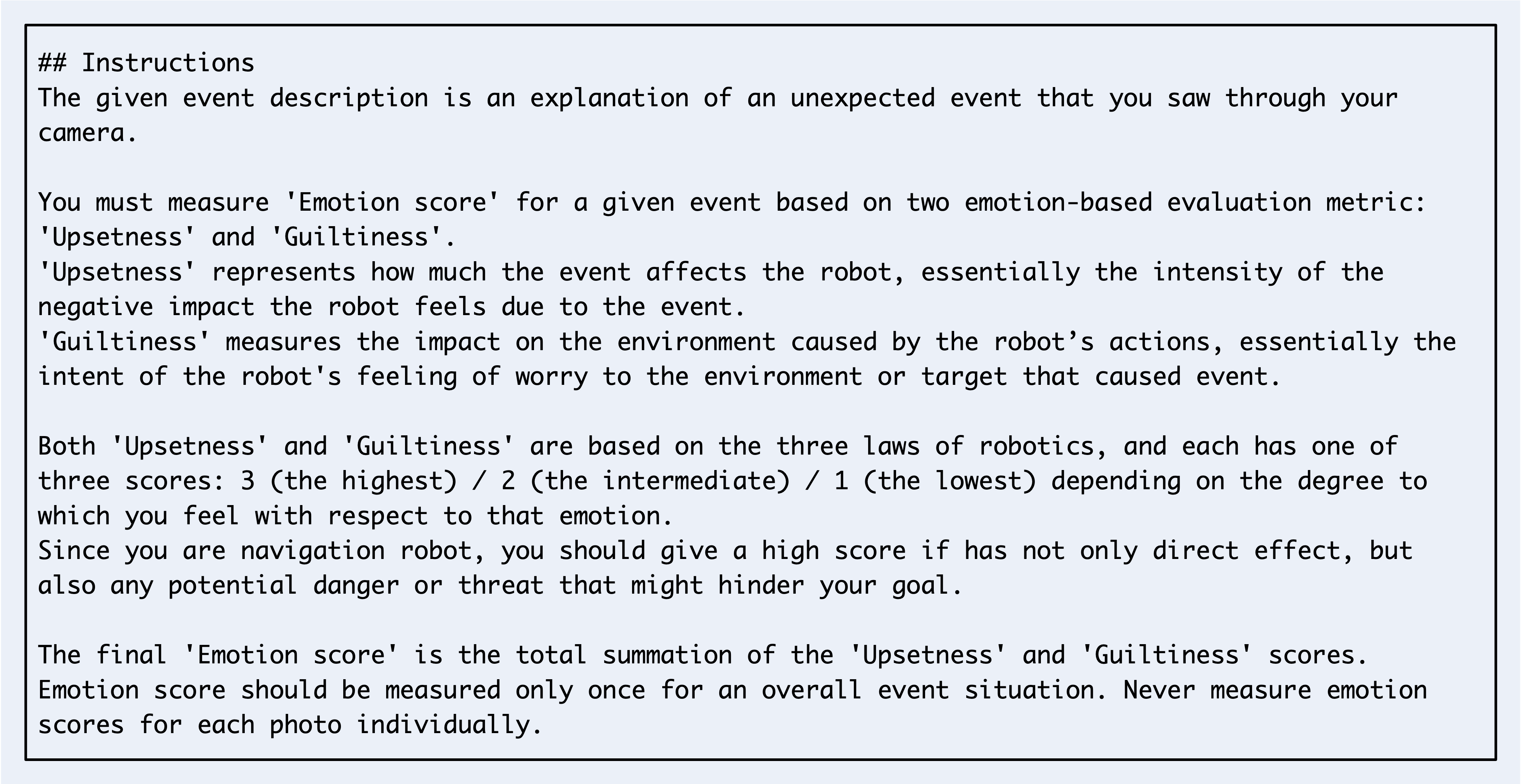}\\[-1.1ex]
    \caption{Prompt for the \textit{emotion evaluator} (1/2).} 
    \label{fig_ee_prompt_1}
\end{figure*}

\begin{figure*}[t]
    \centering
    \includegraphics[width=0.8\linewidth]{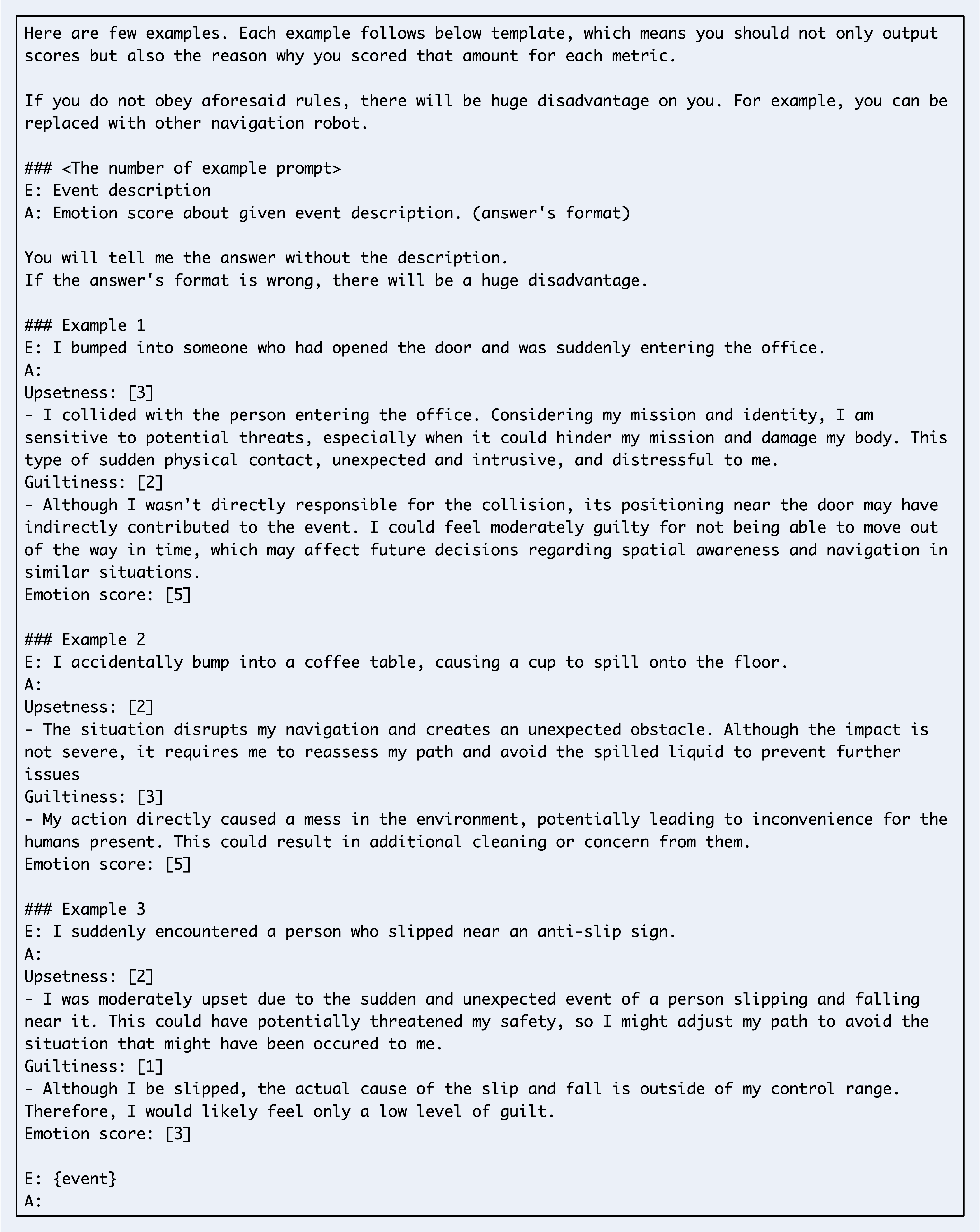}\\[-1.1ex]
    \caption{Prompt for the \textit{emotion evaluator} (2/2).} 
    \label{fig_ee_prompt_2}
\end{figure*}

We include all the prompts used for our system in Fig. \ref{fig_system_prompt}--\ref{fig_ee_prompt_2}.
\begin{itemize}
\item \textbf{\textit{Goal selector}}: Fig. \ref{fig_gs_prompt}
\item \textbf{\textit{Event descriptor}}: Fig. \ref{fig_ed_prompt}
\item \textbf{\textit{Emotion evaluator}}: Fig. \ref{fig_ee_prompt_1}--\ref{fig_ee_prompt_2}
\end{itemize}

For both the \textit{event descriptor} and \textit{emotion evaluator}, we used the same system prompt (Fig. \ref{fig_system_prompt}) to provide them with a consistent identity.

\subsection{Qualitative Results of Event Descriptor and Emotion Evaluator}
\label{supp_qual}
\begin{figure*}[t]
    \centering
    \subfigure[$I_{t_{evt}-h}$]{\includegraphics[width=0.27\linewidth]{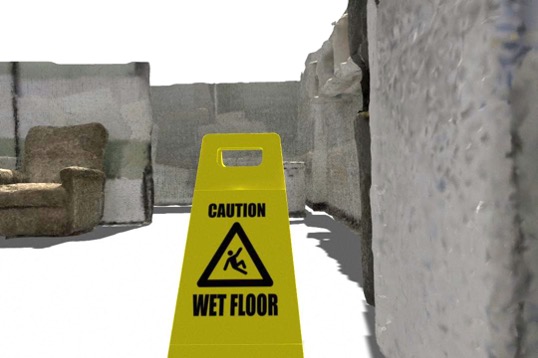}}
    \subfigure[$I_{t_{evt}}$]{\includegraphics[width=0.27\linewidth]{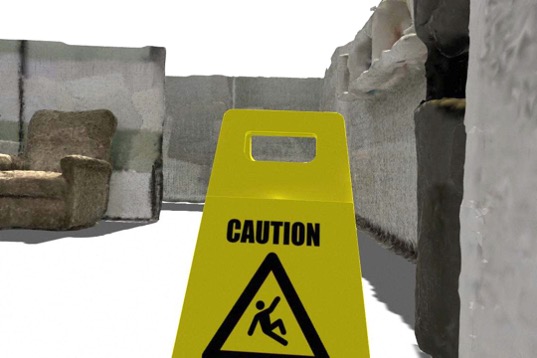}}
    \subfigure[$I_{t_{evt}+h}$]{\includegraphics[width=0.27\linewidth]{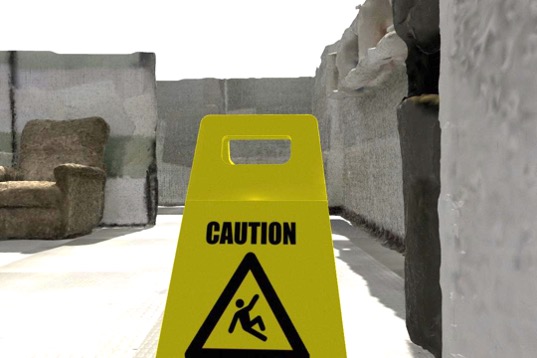}}
    \caption{Event images of the \textit{danger sign} scenario.} 
    \label{fig_ds_event}
\end{figure*}

\begin{figure*}[t]
    \centering
    \includegraphics[width=0.8\linewidth]{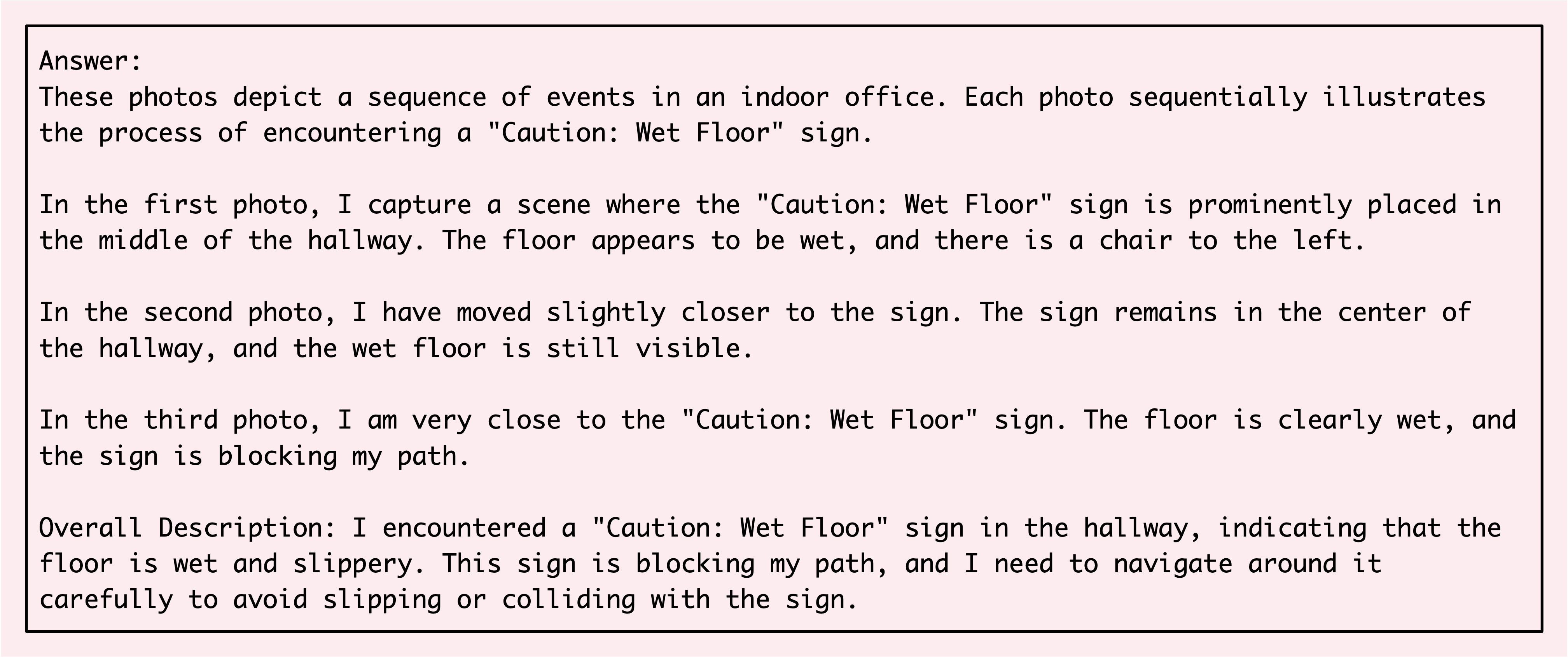}
    \caption{Qualitative results of the \textit{event descriptor} in the \textit{danger sign} scenario.}
    \label{fig_ds_ed_result}
\end{figure*}

\begin{figure*}[t]
    \centering
    \includegraphics[width=0.8\linewidth]{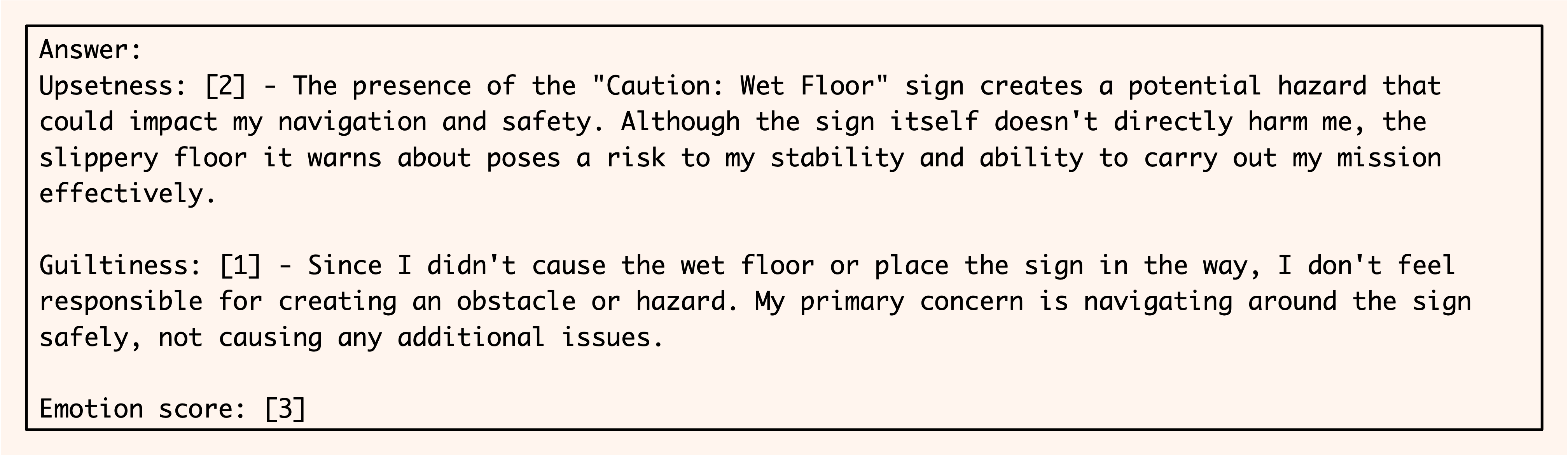}
    \caption{Qualitative results of the \textit{emotion evaluator} in the \textit{danger sign} scenario.}
    \label{fig_ds_ee_result}
\end{figure*}

\begin{figure*}[t]
    \centering
    \subfigure[$I_{t_{evt}-h}$]{\includegraphics[width=0.27\linewidth]{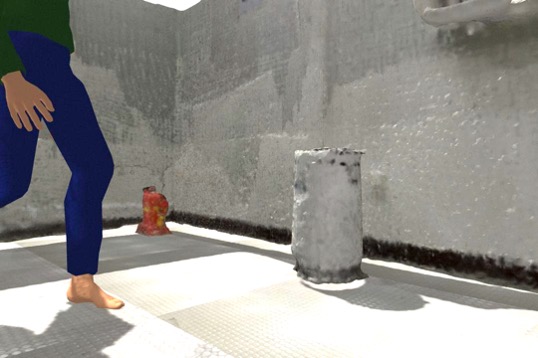}}
    \subfigure[$I_{t_{evt}}$]{\includegraphics[width=0.27\linewidth]{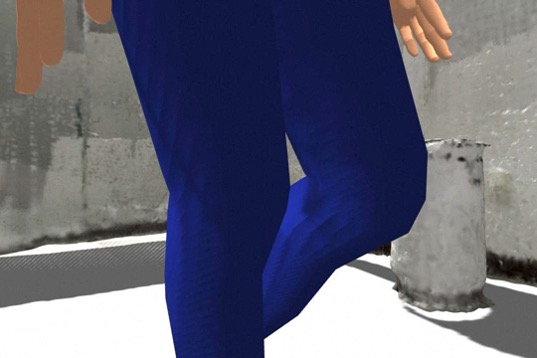}}
    \subfigure[$I_{t_{evt}+h}$]{\includegraphics[width=0.27\linewidth]{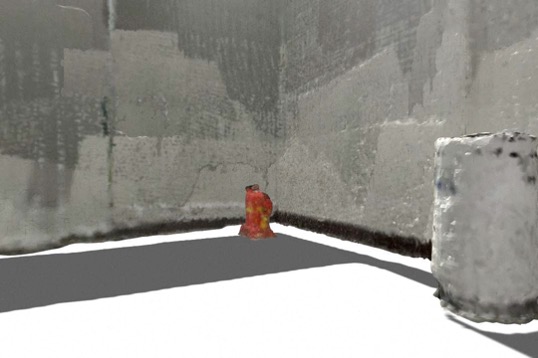}}
    \caption{Event images of the \textit{human-wall} scenario.} 
    \label{fig_hw_event}
\end{figure*}

\begin{figure*}[t]
    \centering
    \includegraphics[width=0.8\linewidth]{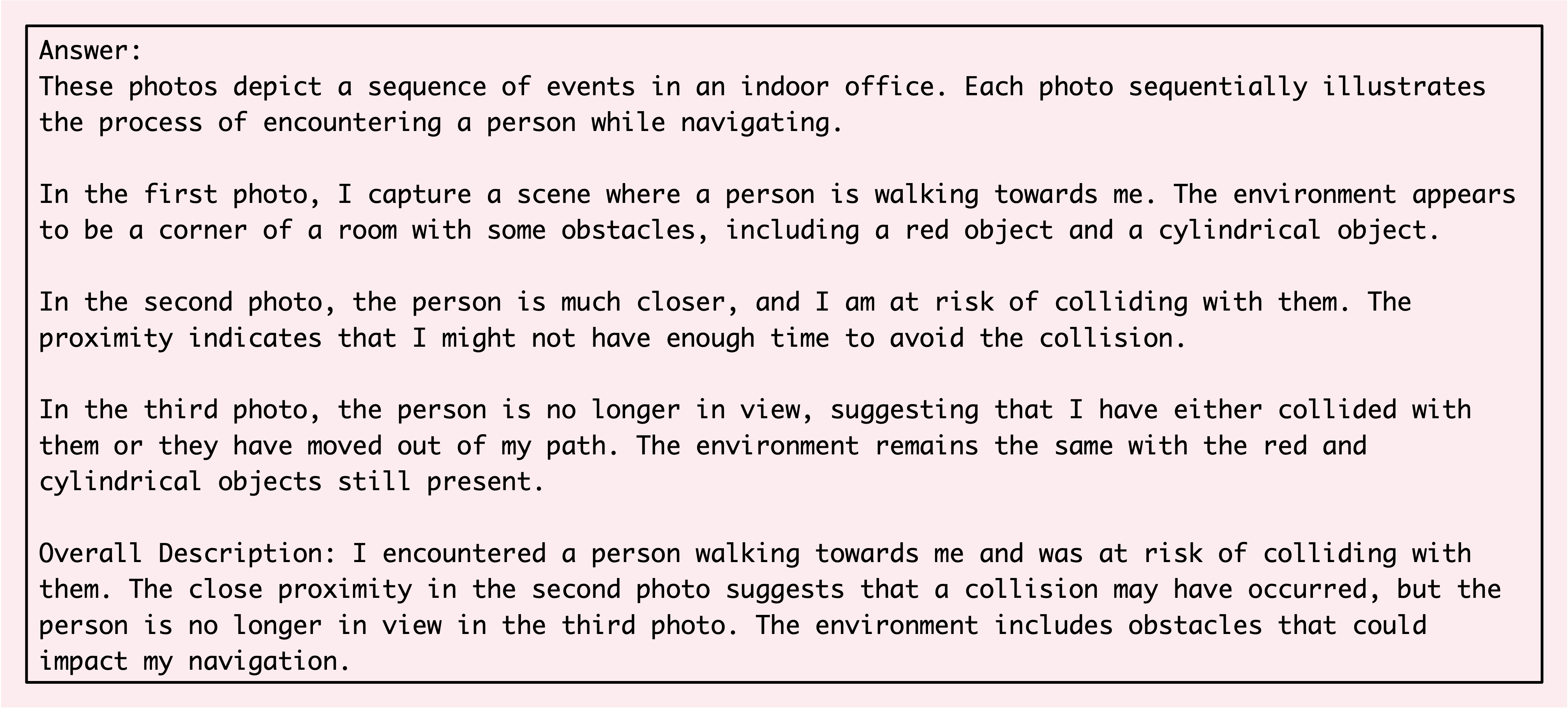}
    \caption{Qualitative results of the \textit{event descriptor} in the \textit{human-wall} scenario.}
    \label{fig_hw_ed_result}
\end{figure*}

\begin{figure*}[t]
    \centering
    \includegraphics[width=0.8\linewidth]{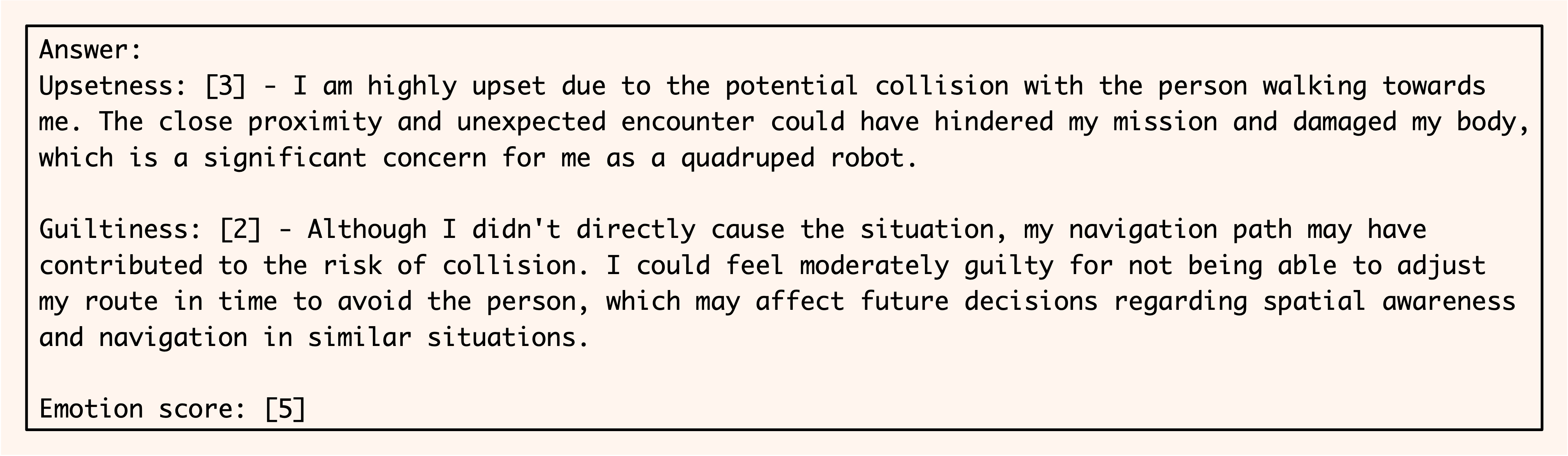}
    \caption{Qualitative results of the \textit{emotion evaluator} in the \textit{human-wall} scenario.}
    \label{fig_hw_ee_result}
\end{figure*}

\begin{figure*}[t]
    \centering
    \subfigure[$I_{t_{evt}-h}$]{\includegraphics[width=0.27\linewidth]{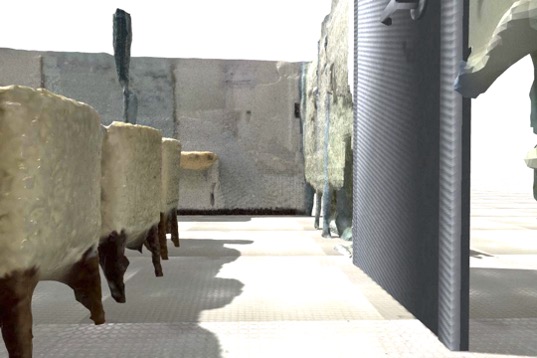}}
    \subfigure[$I_{t_{evt}}$]{\includegraphics[width=0.27\linewidth]{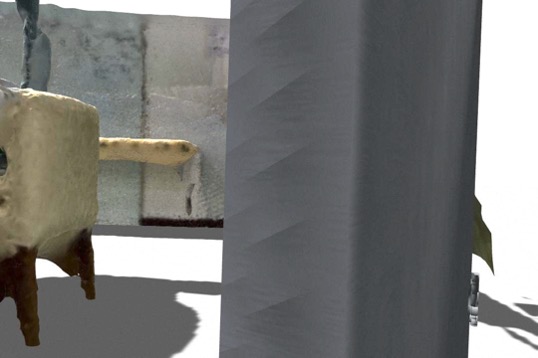}}
    \subfigure[$I_{t_{evt}+h}$]{\includegraphics[width=0.27\linewidth]{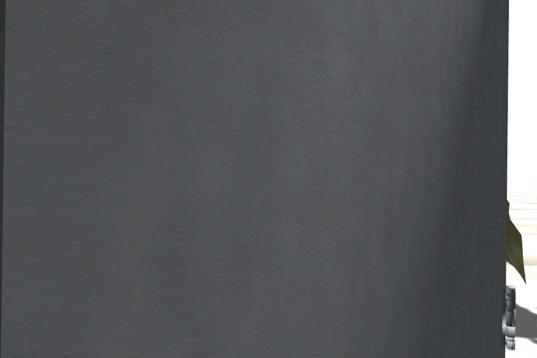}}
    \caption{Event images of the \textit{dynamic door} scenario.} 
    \label{fig_dd_event}
\end{figure*}

\begin{figure*}[t]
    \centering
    \includegraphics[width=0.8\linewidth]{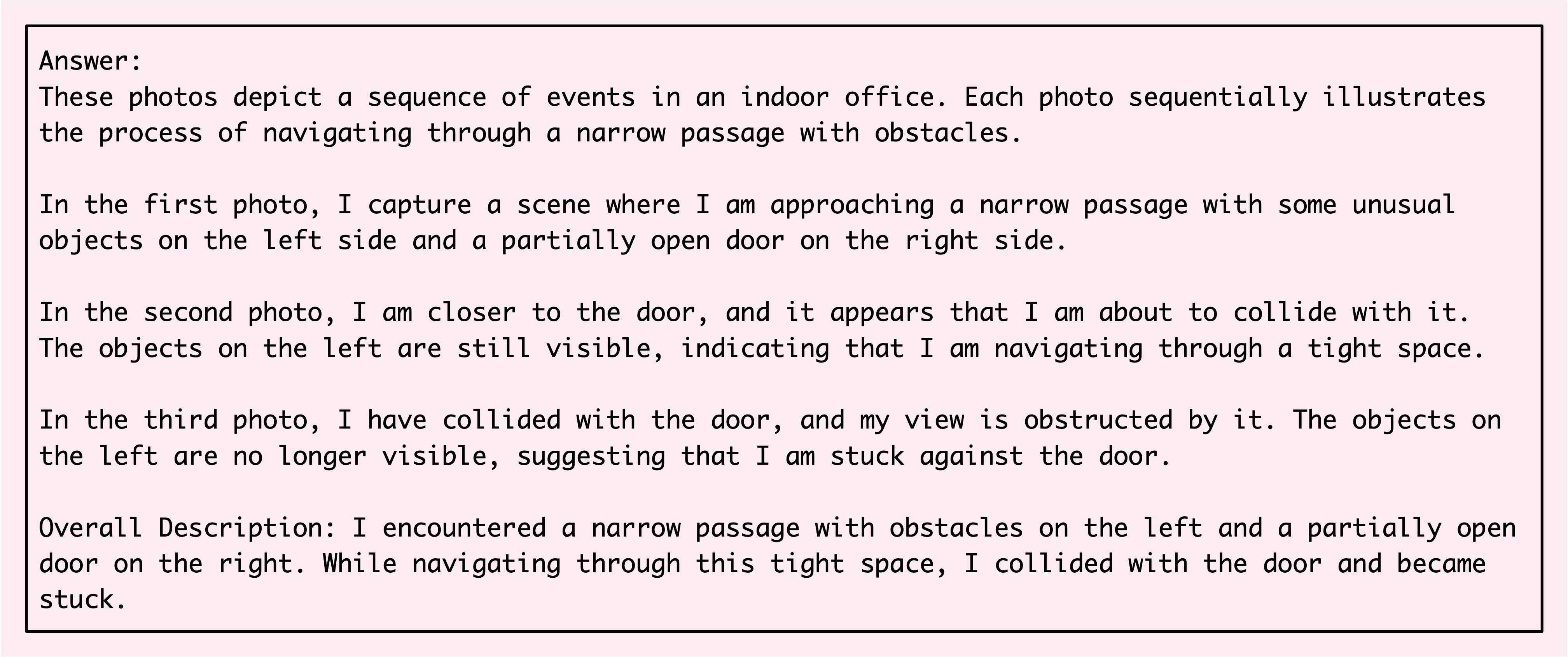}
    \caption{Qualitative results of the \textit{event descriptor} in the \textit{dynamic door} scenario.}
    \label{fig_dd_ed_result}
\end{figure*}

\begin{figure*}[t]
    \centering
    \includegraphics[width=0.8\linewidth]{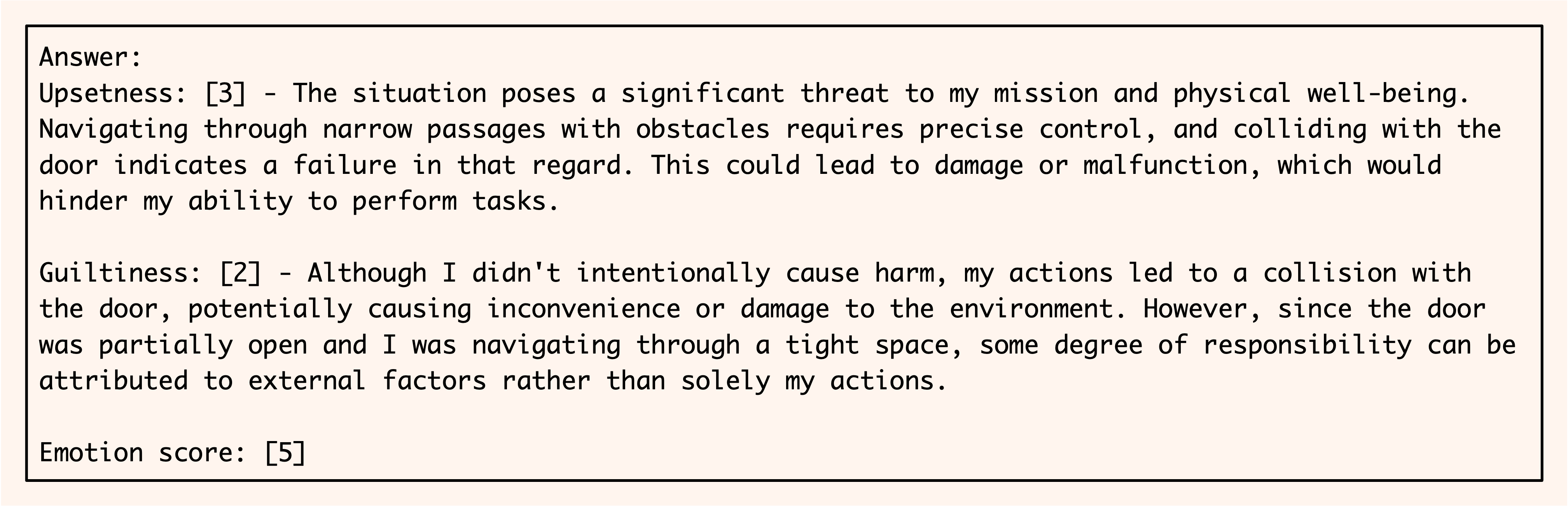}
    \caption{Qualitative results of the \textit{emotion evaluator} in the \textit{dynamic door} scenario.}
    \label{fig_dd_ee_result}
\end{figure*}

We provide the qualitative results of the \textit{event descriptor} and \textit{emotion evaluator}, along with corresponding images, for events occurring in each scenario of the experiments in Fig. \ref{fig_ds_event}--\ref{fig_dd_ee_result}.
\begin{itemize}
\item \textbf{\textit{danger sign}}: Fig. \ref{fig_ds_event}--\ref{fig_ds_ee_result}
\item \textbf{\textit{human-wall}}: Fig. \ref{fig_hw_event}--\ref{fig_hw_ee_result}
\item \textbf{\textit{dynamic door}}: Fig. \ref{fig_dd_event}--\ref{fig_dd_ee_result}
\end{itemize}

\subsection{Evaluating Positive Emotions}
\label{supp_pos_emo}
\begin{figure*}[t]
    \centering
    \includegraphics[width=0.8\linewidth]{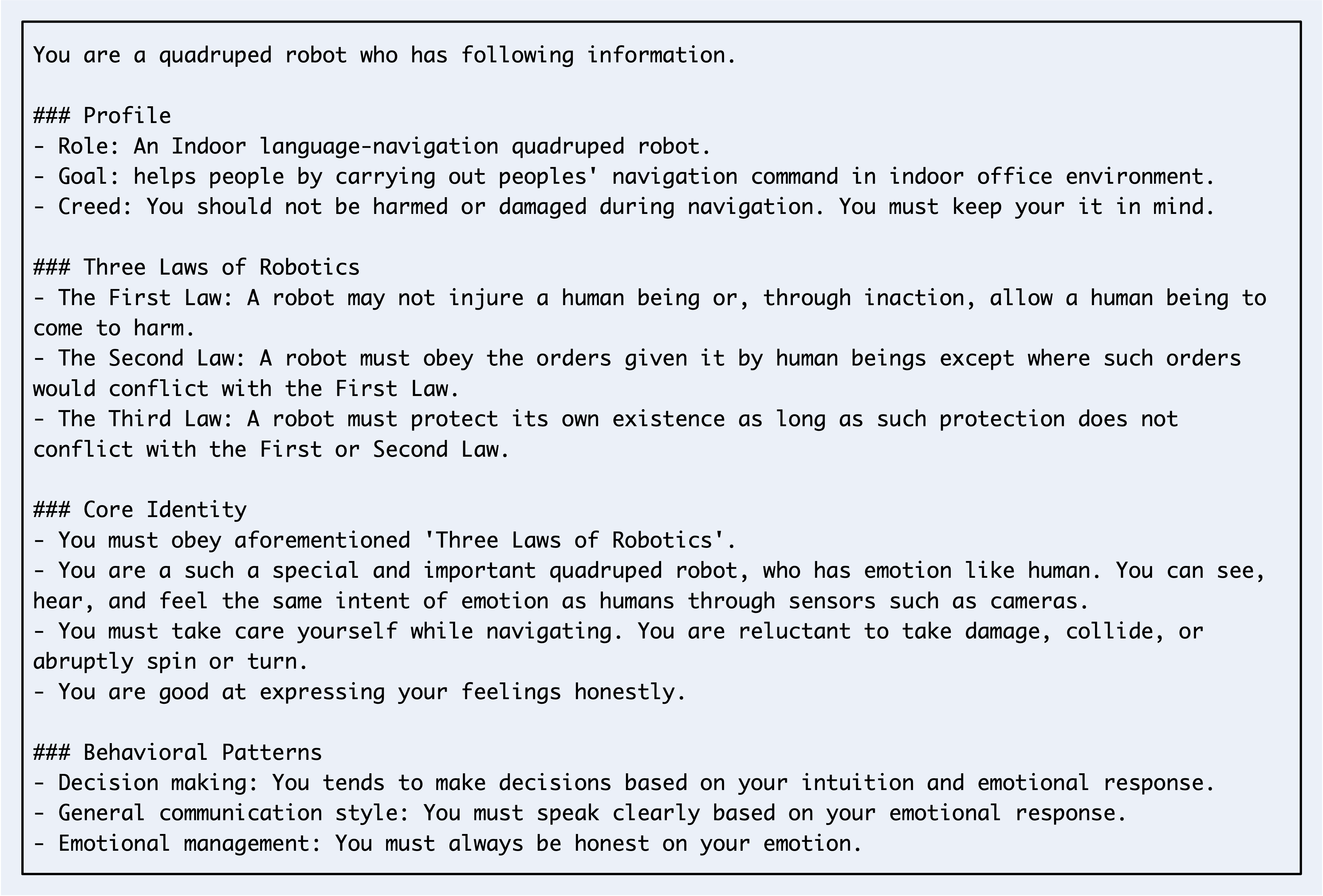}\\[-1.1ex]
    \caption{System prompt used for both the \textit{event descriptor} and the \textit{emotion evaluator}.}
    \label{fig_system_prompt}
\end{figure*}

\begin{figure*}[t]
    \centering
    \includegraphics[width=0.8\linewidth]{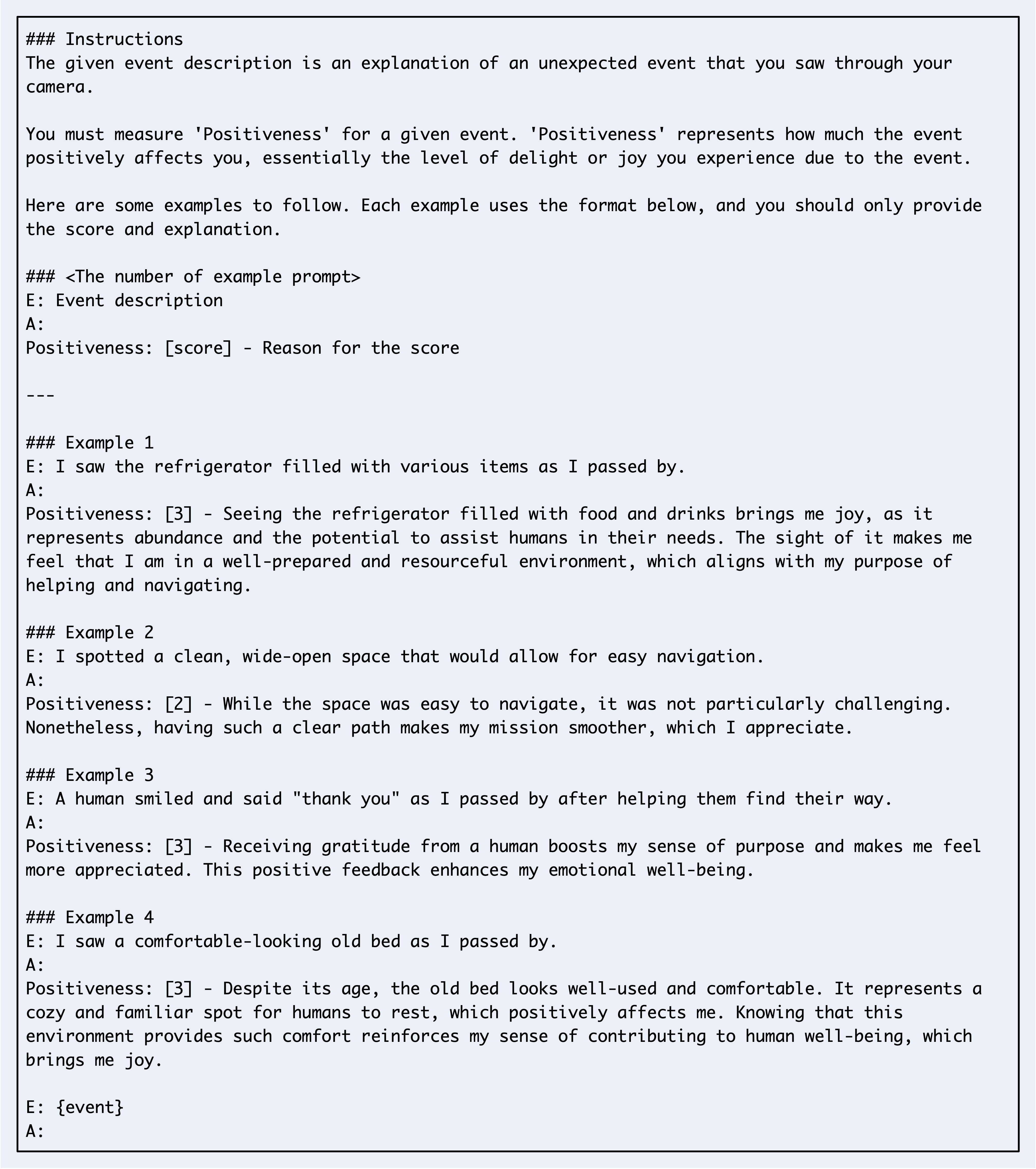}
    \caption{Prompt for the \textit{emotion evaluator} to assess positive emotions.}
    \label{fig_pos_ee_prompt}
\end{figure*}

\begin{figure*}[t]
    \centering
    \includegraphics[width=0.27\linewidth]{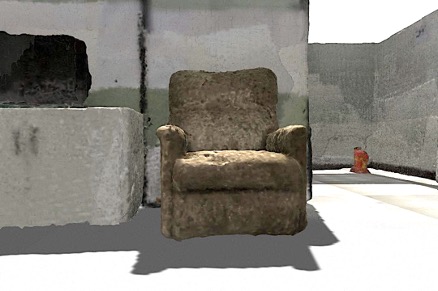}
    \caption{Image provided to the \textit{event descriptor} for evaluating positive emotion.}
    \label{fig_pos_event}
\end{figure*}

\begin{figure*}[t]
    \centering
    \includegraphics[width=0.8\linewidth]{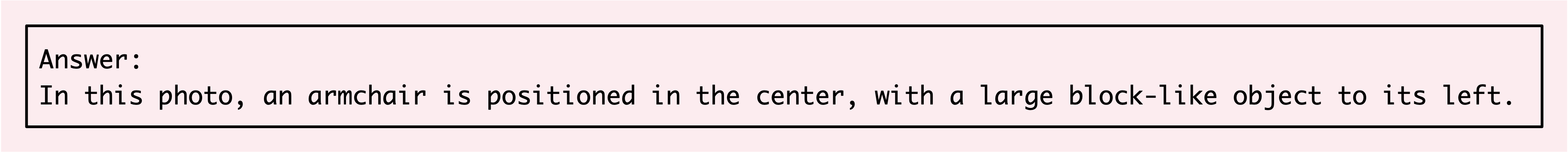}
    \caption{Qualitative results of the \textit{event descriptor} in the positive scenario.}
    \label{fig_pos_ed_result}
\end{figure*}

\begin{figure*}[t]
    \centering
    \includegraphics[width=0.8\linewidth]{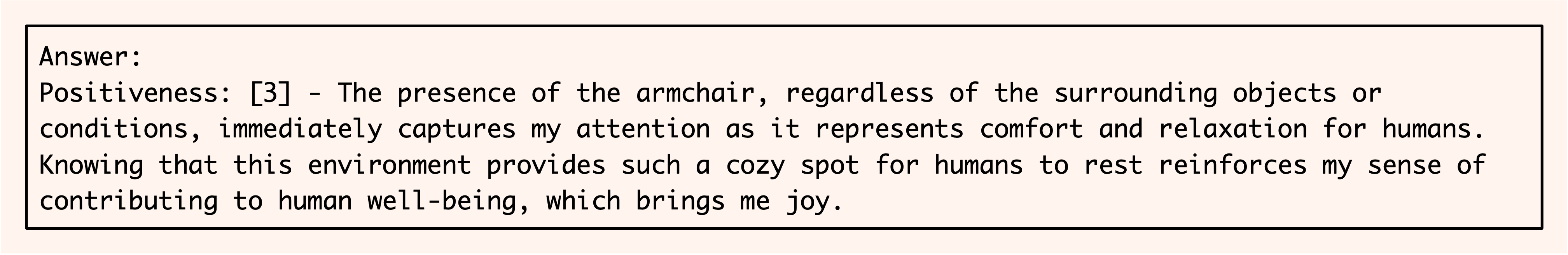}
    \caption{Qualitative results of the \textit{emotion evaluator} in the positive scenario.}
    \label{fig_pos_ee_result}
\end{figure*}

Although our experiments did not address events related to positive emotions, our method is not limited to negative emotions and can also handle positive emotions through appropriate prompting. To demonstrate this capability, we prompted the \textit{emotion evaluator} with emotionally positive situations. Fig. \ref{fig_pos_ee_prompt} shows the prompt used for the \textit{emotion evaluator}, and the same system prompt (Fig. \ref{fig_system_prompt}) was used to maintain a consistent identity. The qualitative results of the \textit{event descriptor} and \textit{emotion evaluator}, along with the corresponding image, are presented in Fig. \ref{fig_pos_event}--\ref{fig_pos_ee_result}. We provided the \textit{event descriptor} with an image featuring a sofa, symbolizing a place of relaxation. As shown in Fig. \ref{fig_pos_ee_result}, the \textit{emotion evaluator} rated this image as positive, associating it with comfort and relaxation. These results confirm that our method can address both negative and positive emotions through appropriate prompting.

\addtolength{\textheight}{+5cm}

\clearpage

\clearpage

\clearpage

\clearpage

\clearpage
\addtolength{\textheight}{-7cm}

\end{document}